\documentclass[sigconf]{acmart}
\AtBeginDocument{%
  }

\setcopyright{acmlicensed}

\acmConference[Conference acronym 'XX]{Make sure to enter the correct
  conference title from your rights confirmation email}{June 03--05,
  2018}{Woodstock, NY}

\usepackage{amsfonts, amsmath, amsthm}
\usepackage[ruled, vlined, linesnumbered]{algorithm2e}
\usepackage{graphicx}
\usepackage{multirow}
\usepackage{booktabs}
\usepackage{makecell}
\usepackage{bm}
\usepackage{array}
\usepackage{tabularx}
\usepackage{nccmath}
\usepackage{subfig}
\usepackage{xcolor, xspace}

\newcommand\figref[1]{Fig.~\ref{#1}}

\newcommand\tabref[1]{Tab.~\ref{#1}}
\newcommand\secref[1]{Sec.~\ref{#1}}

\newcommand{\fakeparagraph}[1]{\par\noindent\textbf{#1.}}

\newcommand{\sysname}{RecursiveECG\xspace}
\begin{document}

\title{Failures Reveal What Metrics Miss: An Evidence-Driven Agent for Recursive Refinement of ECG Classifiers}


\author{Jinliang Deng}
\affiliation{%
  \institution{Beihang University}
  \department{School of Computer Science and Engineering}
  \city{Beijing}
  \country{China}
}
\email{xxx@xxx.com}

\author{Yiming Niu}
\affiliation{%
  \institution{Beihang University}
  \department{School of Computer Science and Engineering}
  \city{Beijing}
  \country{China}
}
\email{xxx@xxx.com}

\author{Yibo Pan}
\affiliation{%
  \institution{Beihang University}
  \department{School of Artificial Intelligence}
  \city{Beijing}
  \country{China}
}
\email{xxx@xxx.com}

\author{Zhiqi Shao}
\affiliation{%
  \institution{Chongqing University}
  \department{School of Economics and Business Administration}
  \city{Chongqing}
  \country{China}
}
\email{zsha2911@uni.sydney.edu.au}

\author{Qin Luo}
\affiliation{%
  \institution{Fuwai Hospital}
  \department{Department of Cardiology}
  \city{Beijing}
  \country{China}
}
\email{luoqin2009@163.com}

\author{Yongxin Tong}
\affiliation{%
  \institution{Beihang University}
  \department{School of Computer Science and Engineering}
  \city{Beijing}
  \country{China}
}
\email{yxtong@buaa.edu.cn}

\makeatletter
\renewcommand{\@mkauthors@iii}{%
  \global\setbox\mktitle@bx=\vbox{%
    \noindent
    \unvbox\mktitle@bx
    \par\medskip

    \centering

    {\Large
      Jinliang Deng\textsuperscript{1},\quad
      Yiming Niu\textsuperscript{1},\quad
      Yibo Pan\textsuperscript{2},\quad
      Zhiqi Shao\textsuperscript{3},\quad
      Qin Luo\textsuperscript{4},\quad
      Yongxin Tong\textsuperscript{1}
      \par
    }

    \vspace{0.65em}

    {\normalsize
      \textsuperscript{1}School of Computer Science and Engineering,
      Beihang University, Beijing, China\\

      \textsuperscript{2}School of Artificial Intelligence,
      Beihang University, Beijing, China\\

      \textsuperscript{3}School of Economics and Business Administration, Chongqing University,
      Chongqing, China\\

      \textsuperscript{4}Department of Cardiology,
      Fuwai Hospital, Beijing, China\\[2pt]

      \href{mailto:xxx@xxx.com}{dengjinliang@buaa.edu.cn};
      \href{mailto:xxx@xxx.com}{yimingniu@buaa.edu.cn};
      \href{mailto:yibo@.com}{pysoible@buaa.edu.cn};
      \href{mailto:zsha2911@uni.sydney.edu.au}
            {zsha2911@uni.sydney.edu.au};
      \href{mailto:luoqin2009@163.com}
            {luoqin2009@163.com};
      \href{mailto:yxtong@buaa.edu.cn}
            {yxtong@buaa.edu.cn}
      \par
    }

    \bigskip
  }%
}
\makeatother

\renewcommand{\shortauthors}{Deng et al.}

\begin{abstract}

Deep models have substantially advanced 12-lead ECG classification, yet their refinement still relies heavily on human experts to inspect failures and iteratively revise classifier designs. Recent LLM-based agents have demonstrated the potential for automated model design, but when guided only by aggregate performance metrics, they lack insight into why individual cases fail and how the classifier should be revised. We present \sysname{}, an evidence-driven LLM-as-Designer framework in which an LLM serves as an offline model designer that refines ECG classifiers based on concrete failures and objective ECG evidence. To ground failure diagnosis in executable evidence, \emph{Criteria-to-Measurement Compilation} converts curated ECG criteria into validated deterministic functions that produce reproducible, reference-backed measurements for individual ECGs. Building on these measurements, \emph{Evidence-Grounded Failure Review} analyzes failed and comparator cases by jointly considering raw waveforms, measurements, and model outputs, enabling the LLM to diagnose classifier limitations and formulate targeted revisions. Candidate revisions are executed and re-evaluated under a fixed problem contract, and only evidence-supported updates are retained. The resulting predictor is frozen after refinement and requires no LLM inference during deployment, while an audit trail links each accepted revision to its supporting evidence. Across PTB-XL, Georgia, and CPSC2018, \sysname{} consistently outperforms strong baselines, achieving an average relative improvement of 10.0\%. Extensive ablation and transfer studies further validate the effectiveness of its evidence-grounded refinement process.
\end{abstract}



\maketitle

\section{Introduction}
\label{sec:introduction}

Electrocardiography remains a core examination for cardiovascular diagnosis because a standard 12-lead recording can characterize cardiac activation and repolarization from multiple electrical perspectives while remaining inexpensive and widely accessible~\cite{kligfield2007ecgpart1}.
However, ECG diagnosis is challenging because it requires models to capture subtle local waveform morphologies and their complex interactions across time, waveform components, and leads~\cite{na2024stmem}.

\begin{figure*}[t]
    \centering
    \includegraphics[width=0.98\textwidth, keepaspectratio]{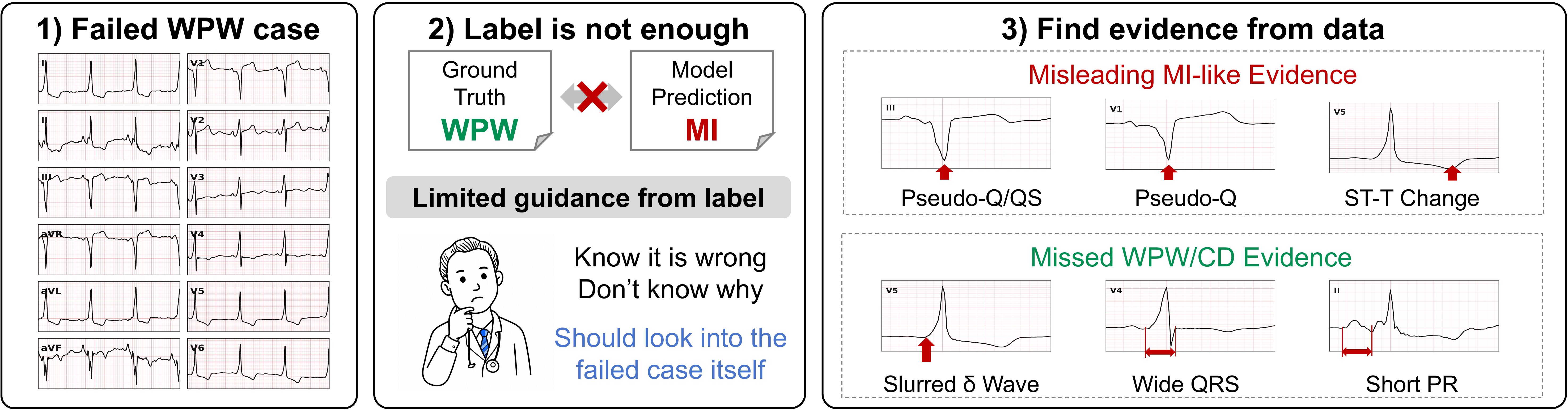}
    \caption{
    Motivation example illustrating how ECG model failures can be analyzed using evidence for refinement.
    }
    \label{fig:wpw_motivation}
\end{figure*}

The development of deep learning models, together with the construction of large-scale benchmark datasets, has substantially advanced ECG classification~\cite{wagner2020ptbxl,strodthoff2021deep,na2024stmem,zhou2025htuning,liu2024merl,wang2025melp,kang2025xlstmecg}.
However, further progress still heavily relies on human-led model refinement, which requires expertise in both medicine and deep learning and often involves substantial time and effort.
To improve upon existing model designs, developers usually need to inspect failure patterns, diagnose model limitations, formulate revision hypotheses, implement corresponding changes, and evaluate them in a closed loop.
This process suggests that advancing ECG classification is not merely a matter of training a stronger classifier, but also a failure-driven model design problem: one must understand where the current classifier fails and how its design should be revised.

A natural question is whether large language models can reduce this refinement burden.
Many existing LLM-based studies on time-series analysis, however, use LLMs in a direct \textit{LLM-as-Classifier} manner, where each time-series sample is fed into an LLM for instance-level prediction or judgment~\cite{jin2024timellm}.
Although flexible, this paradigm requires per-record LLM inference, incurs high computational cost, and is difficult to reproduce reliably due to hallucination and signal-to-text information loss. 

Inspired by recent agentic discovery frameworks such as AlphaEvolve \cite{novikov2025alphaevolve} and LLM-SR \cite{shojaee2025llmsr}, we adopt an \textit{LLM-as-Designer} paradigm for ECG classification, where the LLM serves as an offline model-design controller rather than an online diagnostic classifier as in the LLM-as-Classifier paradigm. 
Driven by feedback from iterative evaluation, the agent reasons about the limitations of the current classifier, proposes executable revisions, and validates the effectiveness of these revisions under a fixed protocol.
The final output is a frozen ECG classifier that can be deployed without LLM inference, making deployment-time prediction more reproducible and traceable.

The key question, however, is what feedback should drive this design process.
In general agentic discovery settings, candidate artifacts are often revised and selected based on aggregate validation metrics, such as accuracy or task-level rewards~\cite{huang2024mlagentbench,novikov2025alphaevolve,ma2024eureka}.
While such metric-driven feedback is effective for judging whether a generated solution performs better overall, it is insufficient for ECG model refinement.
Specifically, aggregate metrics provide limited guidance on why a particular ECG case fails, which signal characteristics or model limitations caused the error, or which diagnostic criteria or modeling mechanisms should be incorporated into the next revision.
This limitation is especially critical for ECG classification, where many errors arise from subtle diagnostic criteria.

As shown in \figref{fig:wpw_motivation}, a CNN-based classifier mistakes a Wolff--Parkinson--White (WPW) case under the conduction disturbance (CD) superclass as myocardial infarction (MI). This error is clinically plausible because WPW-related pre-excitation can distort the QRS complex and ST--T segment, producing patterns resembling ECG manifestations of MI. Yet the same ECG contains conduction-related evidence supporting the CD label, including a short PR interval, a slurred delta wave, and a widened QRS complex. 
The label mismatch exposes the error but not why the classifier favored misleading MI-like cues or how to correct this error. 
A useful repair signal must ground failures in measurable, reference-backed ECG evidence and translate attribution into concrete classifier revisions.

Building on this insight, we introduce \sysname, an evidence-driven LLM-as-Designer framework in which the LLM serves as an offline controller for ECG classifier refinement. 
\sysname first establishes a problem contract that specifies the label space, data split, evaluation protocol, and leakage constraints and holds them fixed throughout the refinement process. 
It then applies \textit{Criteria-to-Measurement Compilation} to convert curated ECG criteria into validated deterministic measurement functions with explicit computation rules, which are executed on raw ECG signals to extract structured, clinically interpretable features for failure analysis. 
Subsequently, it recursively proposes and evaluates candidate classifiers. The evaluation results are used to select informative failure cases for \textit{Evidence-Grounded Failure Review}, which integrates raw waveforms, reference-backed measurements, model outputs, and comparator cases to identify the underlying classifier weakness and formulate a targeted revision. 
Each revision is then executed and re-evaluated, and only updates supported by case-level evidence are retained.
The process produces a frozen ECG classifier together with an audit trail linking each revision to the failure evidence that motivated it.

\begin{figure}[t]
    \centering
    \includegraphics[width=\columnwidth]{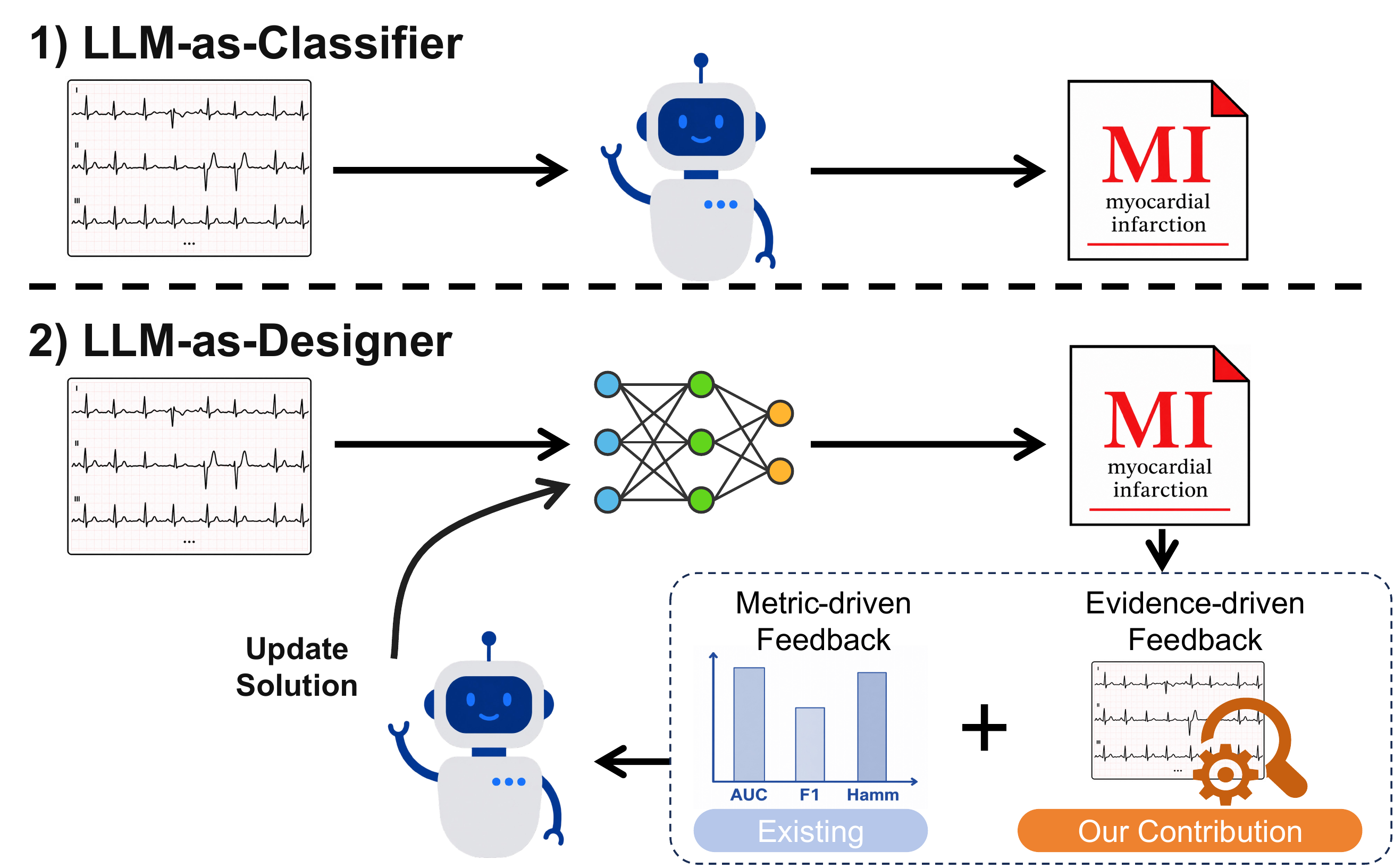}
    \caption{
    Comparison of agentic paradigms for ECG analysis.
    }
    \label{fig:para_comparison}
\end{figure}

\fakeparagraph{Contributions}
In summary, we make the following contributions:

\begin{itemize}
\item We formulate ECG model development as an evidence-driven LLM-as-Designer, where an LLM acts as an offline controller that uses failure evidence to recursively refine a deployable and traceable classifier.

\item We introduce \emph{Criteria-to-Measurement Compilation}, which converts curated ECG criteria into validated deterministic functions that produce reproducible, reference-backed measurements on individual ECGs.

\item We introduce \emph{Evidence-Grounded Failure Review}, which combines failed and comparator cases, waveform measurements, and model outputs to attribute errors and derive targeted, evidence-supported revisions.

\item Across PTB-XL, Georgia, and CPSC2018, \sysname consistently outperforms the strongest baselines, achieving an average relative macro-F1 improvement of 10.0\%, while ablation and transfer studies validate its core mechanisms.
\end{itemize}
\section{Related Work}
\label{sec:related}

\fakeparagraph{ECG diagnostic classification}
Public ECG benchmarks have provided the empirical foundation for modern diagnostic classification. PTB-XL offers large-scale 12-lead ECG waveforms with structured multi-label annotations~\cite{wagner2020ptbxl}, and subsequent studies established strong baselines for diagnostic prediction~\cite{strodthoff2021deep,hannun2019cardiologist,ribeiro2020automatic,attia2019screening,zhu2020automatic,golany2019pgans,golany2020improving}. 
Recent studies improve ECG representations through spatio-temporal masked or contrastive modeling~\cite{na2024stmem,kiyasseh2021clocs}, parameter-efficient adaptation of pre-trained ECG models~\cite{zhou2025htuning}, and specialized sequence architectures such as xLSTM-ECG~\cite{kang2025xlstmecg}. 
ECG-language and foundation-model studies further extend ECG analysis to question answering, report generation, and multimodal pretraining~\cite{oh2023ecgqa,wan2025meit,liu2024merl,wang2025melp}, while clinically guided localization improves interpretability. 
Whereas these studies primarily focus on prediction and representation learning, \sysname focuses on auditable revision of an ECG classifier based on deterministic measurements and case-level failure evidence.

\fakeparagraph{Agents for time-series reasoning}
Existing LLM-based time-series methods can be categorized by the role assigned to the language model. 
First, \emph{direct-inference methods} convert numerical observations into LLM-compatible inputs and then use an LLM for reasoning~\cite{gruver2023large,xu2026visualtime,jin2024timellm,zhou2023onefitsall,wang2025chattime}. 
This paradigm offers flexibility at inference time, but its performance depends strongly on the chosen input representation and prompting strategy~\cite{park2025delving}. 
Second, \emph{agentic orchestration methods} use LLMs to generate executable detection rules or iteratively invoke statistical tools~\cite{gu2025argos,yang2025adagent,tao2026anomamind,yao2023react,schick2023toolformer,shinn2023reflexion}. 
These systems coordinate reasoning with
tools, but their adaptability is largely bounded by the capabilities of the available tools and predefined workflows. 
This limits their ability to adapt when the existing toolkit does not adequately capture the domain criteria or failure modes required by the task. 
\sysname addresses this ECG model-revision setting by using an agent to review predictor failures with knowledge-grounded tools, while ensuring that the deployed pipeline requires no LLM inference.

\fakeparagraph{Agents for scientific discovery}
Recent work increasingly uses LLM agents to generate, evaluate, and iteratively revise scientific artifacts~\cite{bran2024chemcrow,sprueill2024chemreasoner,yang2024opro,ma2024eureka,gottweis2026coscientist}. MLAgentBench studies whether agents can inspect experimental results and modify machine-learning code to improve model performance, while Scientific Generative Agent combines LLM-generated scientific structures with differentiable simulation and numerical optimization~\cite{huang2024mlagentbench,ma2024llmsimulation}. LLM-SR applies a related closed-loop paradigm to symbolic regression, using an LLM to propose executable equations and refining them through data-driven evaluation and evolutionary search~\cite{shojaee2025llmsr}. AlphaEvolve generalizes this approach to algorithm discovery by combining LLM-based program mutation, automated evaluators, and evolutionary selection~\cite{novikov2025alphaevolve,romeraparedes2024funsearch}. Beyond candidate optimization, POPPER emphasizes hypothesis falsification by translating scientific hypotheses into measurable implications and testing them through sequential experiments~\cite{huang2025popper}, whereas DiscoveryWorld evaluates whether agents can complete scientific cycles involving hypothesis formation, experimentation, and explanatory inference~\cite{jansen2024discoveryworld}. 
These systems typically guide revision using aggregate objectives and hypothesis-level test results. 
\sysname instead uses case-level waveform evidence, clinical measurements, comparator cases, and predictor outputs to guide revisions of an ECG analysis pipeline.

\section{Problem Formulation}
\label{sec:problem}

\subsection{ECG Classification}

Given a collection of ECG recordings, the goal of ECG classification is to learn a classifier that maps raw ECG signals to diagnostic labels.
Formally, the $i$-th ECG recording is represented as
$
X_i \in \mathbb{R}^{C\times T},
$
where $C$ denotes the number of ECG leads and $T$ denotes the number of temporal samples.
Because multiple diagnostic conditions may coexist in a single recording, its label is represented as a multi-hot vector
$
\mathbf{y}_i\in\mathcal{Y}=\{0,1\}^{L},
$
where $L$ denotes the number of diagnostic labels. An ECG classifier parameterized by $\theta$ is defined as
\[
f_\theta:\mathcal{X}\rightarrow[0,1]^L.
\]
Given a recording $X_i$, the classifier outputs label-wise probabilities
\[
\widehat{\mathbf{p}}_i=f_\theta(X_i),
\]
where $\widehat{p}_{i,l}$ indicates the predicted probability that the $l$-th diagnostic condition is present.
The corresponding binary predictions are obtained using a fixed decision rule:
\[
\widehat{\mathbf{y}}_i
=
\operatorname{Decide}
\left(
\widehat{\mathbf{p}}_i;\boldsymbol{\nu}
\right),
\]
where $\boldsymbol{\nu}$ denotes the predefined label-wise decision thresholds.

The dataset is divided into pairwise disjoint training, validation, and held-out test sets:
$
\mathcal{D}
=
\mathcal{D}_{\mathrm{tr}}
\mathbin{\dot{\cup}}
\mathcal{D}_{\mathrm{val}}
\mathbin{\dot{\cup}}
\mathcal{D}_{\mathrm{te}}.
$
The training set is used to optimize model parameters, whereas the validation set provides feedback during classifier development and refinement.
The held-out test set remains inaccessible throughout refinement and is used only once for final evaluation.

\subsection{ECG Classifier Refinement}

Conventional ECG classification focuses on learning model parameters from training data.
In contrast, we consider the problem of automatically refining an existing ECG classifier through executable modifications to its design and implementation.

Given a trained classifier $f_\theta$, a revision operation $r\in\mathcal{R}$ produces an updated classifier
\[
f_{\theta'}=r(f_\theta),
\]
where $\mathcal{R}$ denotes the space of admissible executable revisions.
The goal of classifier refinement is to identify a revision that addresses limitations of the current classifier while improving predictive performance:
\[
r^{*}
=
\arg\max_{r\in\mathcal{R}}
M\left(r(f_\theta),\mathcal{D}_{\mathrm{val}}\right),
\]
where $M(\cdot)$ denotes the predefined evaluation metric.

While validation performance provides a necessary criterion for comparing candidate revisions, it does not explain why the current classifier fails or what should be modified. Therefore, unlike conventional search that explores predefined model spaces primarily guided by validation metrics, RecursiveECG generates revision hypotheses from case-level failure evidence. Validation performance is used to verify and select evidence-supported revisions, while the held-out test set remains inaccessible throughout refinement and is used exclusively for final evaluation.

\section{Method}
\label{sec:method}

\begin{figure*}[t]
    \centering
    \includegraphics[width=\textwidth]{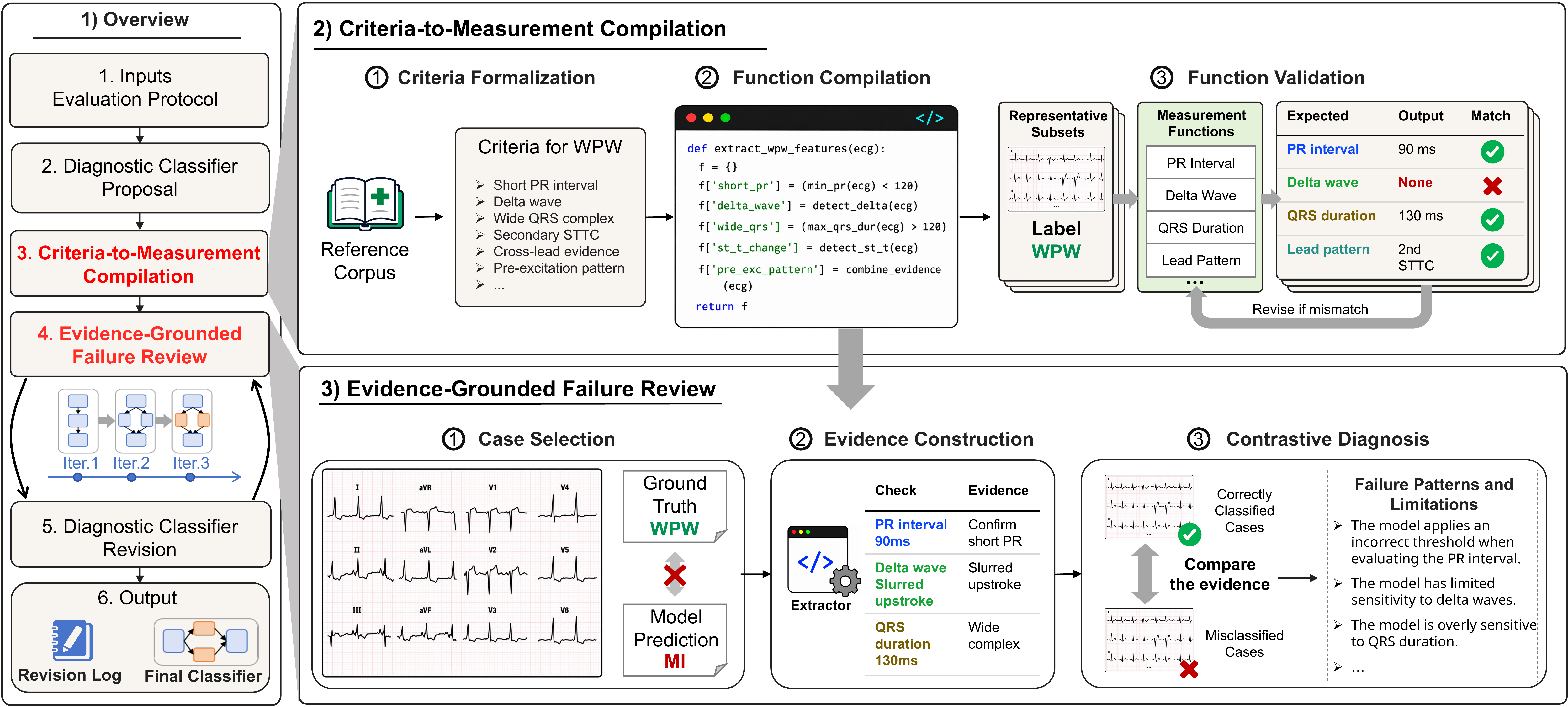}
    \Description{
    Overview of the \sysname workflow.
    Expert-defined ECG criteria are compiled into validated measurement functions.
    The current classifier is evaluated to construct a targeted failure batch, case-level evidence is assembled, and failed cases are contrastively reviewed against correctly classified reference cases.
    The resulting classifier limitations guide LLM-based code revision, followed by execution and validation under a fixed protocol.
    }
    \caption{
    Overview of \sysname.
    It compiles ECG criteria into validated measurement functions, selects influential failure cases, constructs evidence, diagnoses classifier limitations through contrastive review, and performs classifier revision.
    }
    \label{fig:method_overview}
\end{figure*}

\subsection{Overview of \sysname}

\sysname is an agentic framework for automatically refining existing ECG classifiers. 
Rather than using an LLM for sample-level ECG prediction, \sysname adopts an \textit{LLM-as-Designer} paradigm, in which an offline LLM agent analyzes classifier failures, directly modifies the classifier implementation, and validates candidate revisions under a fixed evaluation protocol.

Given an initial ECG classifier, \sysname performs refinement through three stages. 
First, \textit{Criteria-to-Measurement Compilation} transforms expert-defined ECG diagnostic criteria into functionally validated executable measurement functions, which provide clinically grounded signal measurements. 
Second, \textit{Evidence-Grounded Failure Review} selects informative failures, constructs case-level evidence, and contrastively analyzes them against correctly classified reference cases to identify limitations of the current classifier. 
Finally, \textit{Diagnostic Classifier Revision} translates the identified limitations into executable modifications and retains only revisions that improve validation performance. 
The accepted classifier then enters the next refinement iteration, while the held-out test set remains inaccessible until final evaluation.

\subsection{Criteria-to-Measurement Compilation}
\label{sec:criteria_measurement}

Aggregate performance metrics provide limited insight into the ECG characteristics associated with prediction failures.
To construct clinically grounded evidence, \sysname introduces \textit{Criteria-to-Measurement Compilation}, which converts expert-defined ECG diagnostic criteria into functionally validated executable measurement functions.
The resulting measurement inventory and validation contract are summarized in \tabref{tab:app_measurement_inventory} and \secref{app:extractor_validation}.

\fakeparagraph{Clinical Criterion Formalization}
Let
$
\mathcal{C}=\{c_q\}_{q=1}^{Q}
$
denote the collection of curated ECG diagnostic criteria.
Each criterion is formalized as a structured specification
\[
c=(d,m,\tau),
\]
where $d$ denotes the diagnostic concept, $m$ denotes the target ECG measurement, and $\tau$ denotes the condition used to interpret the measurement.

For example, the criterion
``a prolonged PR interval indicates first-degree atrioventricular block''
can be formalized as
\[
c=
\left(
\text{first-degree AV block},
\text{PR interval},
\text{PR}>200\text{ ms}
\right).
\]
This formalization specifies which clinically meaningful property should be extracted and how its value should be interpreted, providing a computable specification for subsequent function compilation.

\fakeparagraph{Measurement Function Compilation and Validation}
Given a formalized criterion $c$, the LLM agent generates an executable measurement function under the compilation instruction $\rho_{\mathrm{cmp}}$:
\[
g_c^{(0)}
=
\mathcal{A}_{\phi}
\left(
c;\rho_{\mathrm{cmp}}
\right).
\]
The key compilation prompt excerpt corresponding to
$\rho_{\mathrm{cmp}}$ is reported in \secref{app:prompt_templates}.
\[
g_c:\mathcal{X}\rightarrow\mathbb{R}^{k_c},
\]
where
$
\mathbf{z}_{i,c}=g_c(X_i)
$
denotes the structured measurements extracted from ECG recording $X_i$, and $k_c$ is the number of values produced for criterion $c$.

Although the function is generated by an LLM, its execution is deterministic: the same ECG input produces the same measurements.
The compiled function therefore operationalizes the clinical criterion as an explicit and reproducible computational procedure rather than an implicit latent representation.
\sysname validates each function on a representative validation subset $\mathcal{V}_c$ covering relevant labels and criterion conditions.
The validation process examines execution correctness and the reliability of the extracted ECG measurements.
At iteration $t$, the resulting feedback is represented as
$\eta_c^{(t)}$.
When a function fails the predefined tests, the feedback is returned to the agent for correction:
\[
g_c^{(t+1)}
=
\mathcal{A}_{\phi}
\left(
g_c^{(t)}, c,\eta_c^{(t)};
\rho_{\mathrm{cmp}}
\right).
\]
This process continues until the generated function satisfies the validation requirements.
The functionally validated measurements form a reusable measurement library
$
\mathcal{G}
=
\{g_c\mid c\in\mathcal{C}\},
$
which provides the clinically grounded measurements used in subsequent failure review.

\subsection{Evidence-Grounded Failure Review}
\label{sec:failure_review}

\sysname introduces \textit{Evidence-Grounded Failure Review}, which transforms failures into actionable evidence through case selection, evidence construction, and batch-level contrastive diagnosis.
Selection and retrieval details are provided in \secref{app:failure_retrieval}.

\fakeparagraph{Failure-driven Case Selection}
Although evaluation may reveal diverse errors, not all failure cases provide equally useful feedback for model refinement.
The agent should therefore prioritize cases that are most likely to expose limitations of the current classifier.

Let the validation failure set of classifier $f_\theta$ be
\[
\mathcal{D}^{-}_{\mathrm{val},\theta}
=
\left\{
(X_i,\mathbf{y}_i)\in\mathcal{D}_{\mathrm{val}}
\;\middle|\;
\widehat{\mathbf{y}}_i\neq\mathbf{y}_i
\right\}.
\]
Rather than uniformly reviewing all cases in $\mathcal{D}^{-}_{\mathrm{val},\theta}$, the agent constructs a targeted failure batch:
\[
\mathcal{B}^{-}_{\theta}\subseteq\mathcal{D}^{-}_{\mathrm{val},\theta}.
\]
The selection process prioritizes diagnostic labels and error modes that exert substantial influence on validation performance, thereby defining a focused set of failures for further review.

\fakeparagraph{Failure Evidence Construction}
For each recording in the selected failure batch, \sysname constructs an evidence representation that connects classifier behavior with clinically interpretable ECG characteristics.
For an ECG recording $X_i$, the clinically grounded measurements are:
\[
Z_i
=
\left\{
\mathbf{z}_{i,c}=g_c(X_i)
\;\middle|\;
c\in\mathcal{C}
\right\}.
\]
The corresponding case-level evidence is defined as:
\[
E_i
=
\left(
X_i,
\mathbf{y}_i,
\widehat{\mathbf{p}}_i,
\widehat{\mathbf{y}}_i,
Z_i
\right),
\]
which contains the raw ECG waveform, ground-truth labels, label-wise prediction probabilities, and criterion-grounded measurements. For a batch $\mathcal{B}$, its evidence representation is:
\[
E(\mathcal{B})
=
\left\{
E_i
\;\middle|\;
(X_i,\mathbf{y}_i)\in\mathcal{B}
\right\}.
\]
In particular, $E(\mathcal{B}^{-}_{\theta})$ provides the signal-level and decision-level context required to diagnose recurring failures in the selected batch.

\fakeparagraph{Contrastive Failure Diagnosis}
A characteristic observed in the failure batch is not necessarily specific to prediction errors, as it may also occur in correctly classified cases. 
To identify systematic limitations of the current classifier, \sysname contrasts the selected failure batch with correctly classified reference cases exhibiting comparable clinical evidence profiles.

Let
$
\mathcal{D}^{+}_{\mathrm{val},\theta}
=
\left\{
(X_j,\mathbf{y}_j)\in\mathcal{D}_{\mathrm{val}}
\;\middle|\;
\widehat{\mathbf{y}}_j=\mathbf{y}_j
\right\}
$
denote the set of validation samples correctly classified by the current classifier.
Given the selected failure batch $\mathcal{B}^{-}_{\theta}$, a reference retrieval procedure constructs
\[
\mathcal{B}^{+}_{\theta}
=
\mathcal{R}
\left(
\mathcal{B}^{-}_{\theta},
\mathcal{D}^{+}_{\mathrm{val},\theta}
\right),
\]
where $\mathcal{R}(\cdot)$ retrieves correctly classified cases whose clinical evidence profiles are comparable to those of the failure batch.

Given the evidence representations of the failure and reference batches, the agent performs batch-level contrastive diagnosis:
\[
\ell
=
\mathcal{A}_{\phi}
\left(
E(\mathcal{B}^{-}_{\theta}),
E(\mathcal{B}^{+}_{\theta});
\rho_{\mathrm{diag}}
\right),
\]
where $\ell$ denotes the identified failure pattern together with the corresponding limitation of the current classifier.
The key failure-review prompt excerpt corresponding to
$\rho_{\mathrm{diag}}$ is reported in \secref{app:prompt_templates}.

By comparing failed and successful cases, the agent identifies signal characteristics or diagnostic patterns that are insufficiently captured by the current classifier.
The diagnosed limitation $\ell$ is then passed to the subsequent classifier revision stage.

\subsection{Diagnostic Classifier Revision}
\label{sec:classifier_revision}

Given the diagnosed limitation $\ell$, \sysname uses the LLM agent to revise the ECG classifier implementation and evaluate the resulting candidate model under a fixed protocol.

\fakeparagraph{Code-level Revision Generation}
Let $P_\theta$ denote the executable implementation of the current classifier $f_\theta$, including its model architecture and associated training procedure.
Conditioned on the diagnosed limitation $\ell$ and the revision instruction $\rho_{\mathrm{rev}}$, the LLM agent generates a code-level revision:
\[
r
=
\mathcal{A}_{\phi}
\left(
P_\theta,
\ell;
\rho_{\mathrm{rev}}
\right).
\]
Here, $r$ denotes an executable modification to the current implementation rather than a natural-language recommendation.
The key prompt excerpt corresponding to $\rho_{\mathrm{rev}}$ is reported in \secref{app:prompt_templates}.

Applying the generated revision produces an updated program:
\[
P_{\theta'}
=
\operatorname{Apply}
\left(
P_\theta,r
\right).
\]
The revised program is then instantiated and trained on the fixed training set to obtain the candidate predictor
$
f_{\theta'}
.
$

\fakeparagraph{Revision Execution and Selection}
LLM-generated code may fail to execute or reduce classifier performance.
Each candidate revision is therefore executed and evaluated under the same training and validation protocol as the current classifier.

A revision is admissible only when its program executes successfully and the resulting classifier satisfies the predefined constraints. Among admissible revisions, a candidate is retained when it improves validation performance:
\[
M
\left(
f_{\theta'},
\mathcal{D}_{\mathrm{val}}
\right)
>
M
\left(
f_{\theta},
\mathcal{D}_{\mathrm{val}}
\right).
\]
An accepted revision replaces the current implementation and initiates the next iteration. Refinement terminates when no admissible improvement is found or when the refinement budget is exhausted.

Because each generated revision is conditioned on the evidence-derived limitation $\ell$, \sysname records the provenance of every accepted update in a refinement history:
\[
\mathcal{H}
\leftarrow
\mathcal{H}
\cup
\left\{
\left(
\ell,
r,
M_{\mathrm{before}},
M_{\mathrm{after}}
\right)
\right\}.
\]
This history links each accepted code modification to its motivating model limitation and observed validation outcome.

\begin{table*}[t]
\centering
\caption{Classification results on PTB-XL, Georgia, and CPSC2018.
Values are reported as mean~$\pm$~standard deviation over five runs.
The best overall results are in bold, and the second-best results are underlined.
The final column reports the relative improvement of the best-performing
\sysname{} variant over the strongest non-\sysname{} baseline.}
\label{tab:main_aggregate}
\scriptsize
\setlength{\tabcolsep}{3pt}
\resizebox{\textwidth}{!}{%
\begin{tabular}{ll|cccc|cc|ccc|c}
\toprule
\multirow{2}{*}{Dataset}
& \multirow{2}{*}{Metric}
& \multicolumn{4}{c|}{Deep Models}
& \multicolumn{2}{c|}{Pretrained Models}
& \multicolumn{3}{c|}{Agentic Models}
& \multirow{2}{*}{Impr (\%)} \\
\cmidrule(lr){3-6}
\cmidrule(lr){7-8}
\cmidrule(lr){9-11}
&
& \makecell[c]{PatchTST}
& \makecell[c]{TimesNet}
& \makecell[c]{TS2Vec}
& \makecell[c]{xLSTM-ECG}
& \makecell[c]{UniTS}
& \makecell[c]{MERL}
& \makecell[c]{Argos}
& \makecell[c]{\sysname\\[-1pt]{\tiny (Qwen3.5-27B)}}
& \makecell[c]{\sysname\\[-1pt]{\tiny (DeepSeek-V4-Pro)}}
& \\
\midrule

\multirow{4}{*}{PTB-XL}
& AUC
& $0.8271{\pm}0.0020$
& $0.8710{\pm}0.0020$
& $0.8683{\pm}0.0015$
& $0.9130{\pm}0.0019$
& $0.8465{\pm}0.0030$
& $0.8964{\pm}0.0007$
& $0.8511{\pm}0.0056$
& $\underline{0.9166{\pm}0.0187}$
& \bm{$0.9373{\pm}0.0066$}
& +2.66\% \\

& Macro F1
& $0.5899{\pm}0.0057$
& $0.6557{\pm}0.0051$
& $0.6566{\pm}0.0022$
& $0.7250{\pm}0.0036$
& $0.6202{\pm}0.0033$
& $0.6912{\pm}0.0019$
& $0.6055{\pm}0.1196$
& $\underline{0.7338{\pm}0.0415}$
& \bm{$0.7653{\pm}0.0088$}
& +5.56\% \\

& Micro F1
& $0.6424{\pm}0.0026$
& $0.6975{\pm}0.0054$
& $0.6903{\pm}0.0017$
& $0.7614{\pm}0.0062$
& $0.6783{\pm}0.0059$
& $0.7394{\pm}0.0045$
& $0.6281{\pm}0.0071$
& $\underline{0.7821{\pm}0.0326}$
& \bm{$0.7930{\pm}0.0039$}
& +4.15\% \\

& Hamming Acc.
& $0.7941{\pm}0.0056$
& $0.8279{\pm}0.0044$
& $0.8229{\pm}0.0017$
& $0.8705{\pm}0.0058$
& $0.8160{\pm}0.0082$
& $0.8561{\pm}0.0096$
& $0.7498{\pm}0.0047$
& \bm{$0.9118{\pm}0.0149$}
& $\underline{0.8929{\pm}0.0177}$
& +4.74\% \\

\midrule

\multirow{4}{*}{Georgia}
& AUC
& $0.8459{\pm}0.0054$
& $0.8522{\pm}0.0047$
& $0.9163{\pm}0.0026$
& $0.9233{\pm}0.0034$
& $0.7149{\pm}0.0029$
& $0.9693{\pm}0.0018$
& $0.8849{\pm}0.0086$
& $\underline{0.9698{\pm}0.0124}$
& \bm{$0.9814{\pm}0.0052$}
& +1.25\% \\

& Macro F1
& $0.5574{\pm}0.0087$
& $0.5579{\pm}0.0091$
& $0.6216{\pm}0.0064$
& $0.6827{\pm}0.0068$
& $0.5511{\pm}0.0075$
& $\underline{0.8130{\pm}0.0046}$
& $0.5911{\pm}0.0064$
& $0.8021{\pm}0.0368$
& \bm{$0.8826{\pm}0.0069$}
& +8.56\% \\

& Micro F1
& $0.5720{\pm}0.0061$
& $0.6325{\pm}0.0073$
& $0.6961{\pm}0.0041$
& $0.7469{\pm}0.0057$
& $0.7567{\pm}0.0052$
& $0.8426{\pm}0.0038$
& $0.6483{\pm}0.0085$
& $\underline{0.8662{\pm}0.0297}$
& \bm{$0.9170{\pm}0.0043$}
& +8.83\% \\

& Hamming Acc.
& $0.8544{\pm}0.0078$
& $0.8750{\pm}0.0060$
& $0.8979{\pm}0.0032$
& $0.9178{\pm}0.0039$
& $0.9199{\pm}0.0048$
& $0.9486{\pm}0.0027$
& $0.8699{\pm}0.0094$
& $\underline{0.9587{\pm}0.0116}$
& \bm{$0.9735{\pm}0.0021$}
& +2.62\% \\

\midrule

\multirow{4}{*}{CPSC2018}
& AUC
& $0.7840{\pm}0.0049$
& $0.7285{\pm}0.0068$
& $0.8390{\pm}0.0031$
& $0.9016{\pm}0.0042$
& $0.8668{\pm}0.0037$
& $0.8047{\pm}0.0029$
& $0.8708{\pm}0.0067$
& $\underline{0.9437{\pm}0.0169}$
& \bm{$0.9746{\pm}0.0036$}
& +8.10\% \\

& Macro F1
& $0.4375{\pm}0.0102$
& $0.4089{\pm}0.0115$
& $0.4935{\pm}0.0076$
& $0.5775{\pm}0.0096$
& $0.5470{\pm}0.0089$
& $0.6931{\pm}0.0063$
& $0.5332{\pm}0.0065$
& $\underline{0.6988{\pm}0.0442}$
& \bm{$0.8027{\pm}0.0078$}
& +15.81\% \\

& Micro F1
& $0.4953{\pm}0.0066$
& $0.4683{\pm}0.0094$
& $0.5988{\pm}0.0059$
& $0.6030{\pm}0.0074$
& $0.5915{\pm}0.0071$
& $\underline{0.7459{\pm}0.0051}$
& $0.6007{\pm}0.0086$
& $0.7204{\pm}0.0395$
& \bm{$0.8352{\pm}0.0056$}
& +11.97\% \\

& Hamming Acc.
& $0.8835{\pm}0.0042$
& $0.8802{\pm}0.0057$
& $0.9051{\pm}0.0028$
& $0.8691{\pm}0.0069$
& $0.8848{\pm}0.0065$
& $\underline{0.9363{\pm}0.0034}$
& $0.9054{\pm}0.0068$
& $0.9199{\pm}0.0173$
& \bm{$0.9613{\pm}0.0030$}
& +2.67\% \\

\bottomrule
\end{tabular}}
\end{table*}

\begin{figure}[!htbp]
\centering
\includegraphics[width=\linewidth]{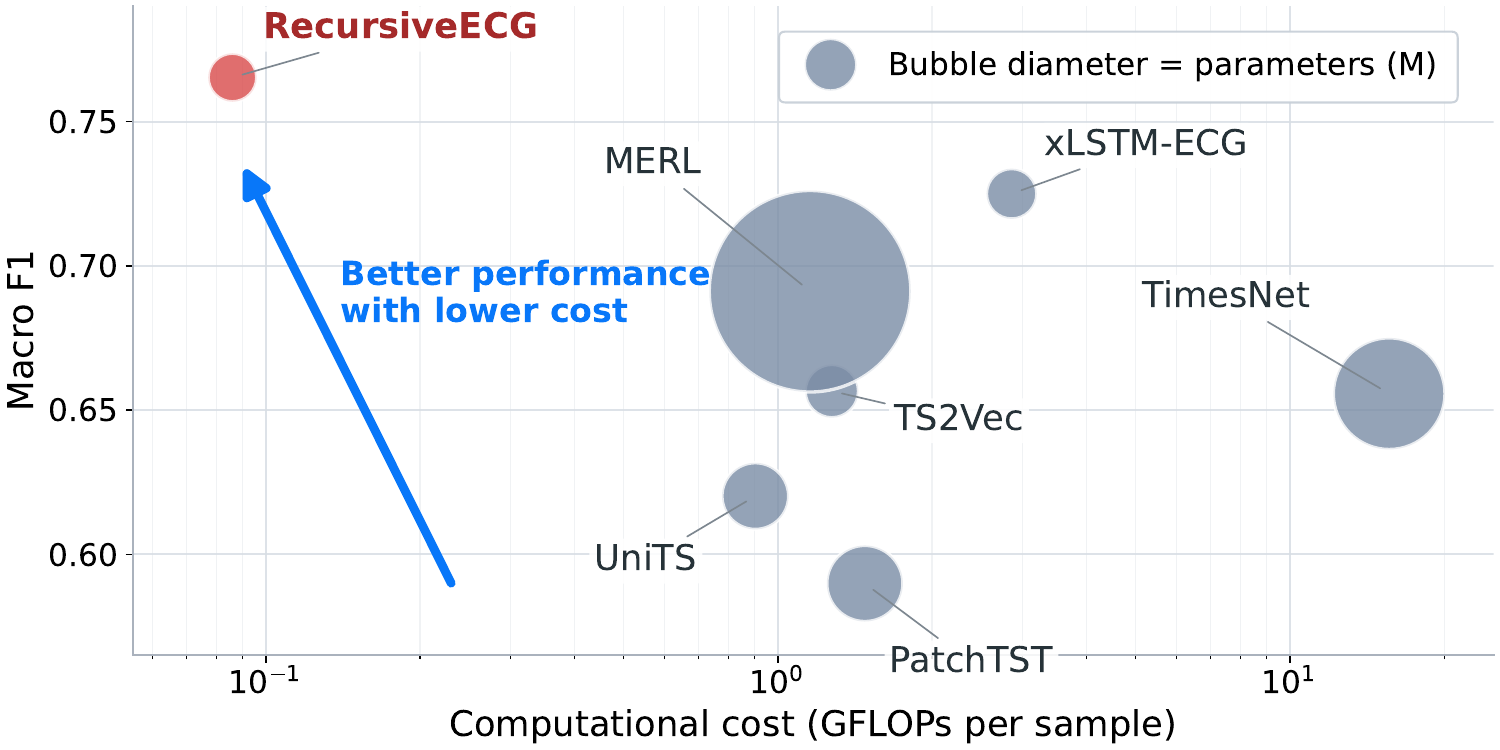}
\caption{The effectiveness--efficiency tradeoff on PTB-XL.}
\label{fig:ptbxl_efficiency_bubble}
\end{figure}

\section{Experiments}
\label{sec:experiments}
In the following section, we conduct a series of experiments to comprehensively evaluate \sysname and demonstrate its effectiveness.
We have made our implementation publicly available at \url{https://github.com/neumyor/RecursiveECG}.

\subsection{Experimental Setup}
\label{sec:datasets}

\fakeparagraph{Datasets}
We use three public 12-lead ECG multi-label diagnostic datasets.
PTB-XL~\cite{wagner2020ptbxl} provides a large-scale diagnostic benchmark with official splits and hierarchical superclass/subclass annotations.
Georgia, drawn from the PhysioNet/Computing in Cardiology Challenge 2020 corpus~\cite{alday2020classification}, evaluates the method on a scored multi-label setting used by prior ECG baselines~\cite{kang2025xlstmecg}.
CPSC2018 provides an additional multi-label arrhythmia benchmark, for which we use a fixed train/validation/test split.
Detailed label definitions and dataset-specific preprocessing choices are provided in \secref{app:protocol}.

\fakeparagraph{Baselines and Protocols}
We compare \sysname{} with both general time-series baselines and ECG-specific baselines, covering diverse paradigms such as deep learning models, pretrained models, and agentic solutions.
Following the order used in the results tables, the deep learning baselines are PatchTST~\cite{nie2023patchtst}, TimesNet~\cite{wu2023timesnet}, TS2Vec~\cite{yue2022ts2vec}, and xLSTM-ECG~\cite{kang2025xlstmecg}; the pretrained or agentic baselines are UniTS~\cite{gao2024units}, MERL~\cite{liu2024merl}, and Argos~\cite{gu2025argos}.
For a fair comparison, all methods use the same data splits and label mapping, with key hyperparameters independently tuned on the validation set of each dataset and model selection performed based on validation performance.
The default implementations of \sysname{} and Argos use DeepSeek-V4-Pro \cite{deepseekai2026deepseekv4} as the LLM agent backbone. For these agentic methods, each run re-executes the entire model design process, allowing different design trajectories, rather than fixing a single architecture and only varying training randomness.
Results are reported as mean~$\pm$~standard deviation over five runs; implementation details are provided in \secref{app:protocol} and \tabref{tab:app_sysname_config}. 

\fakeparagraph{Evaluation Metrics}
Since all three tasks are imbalanced multi-label classification problems, we use Macro F1, Macro AUC, Micro F1, and Hamming accuracy.
For deployment efficiency, we evaluate FLOPs, GPU memory usage, and per-sample inference latency.

\begin{figure}[tbp]
\centering
\includegraphics[width=\linewidth]{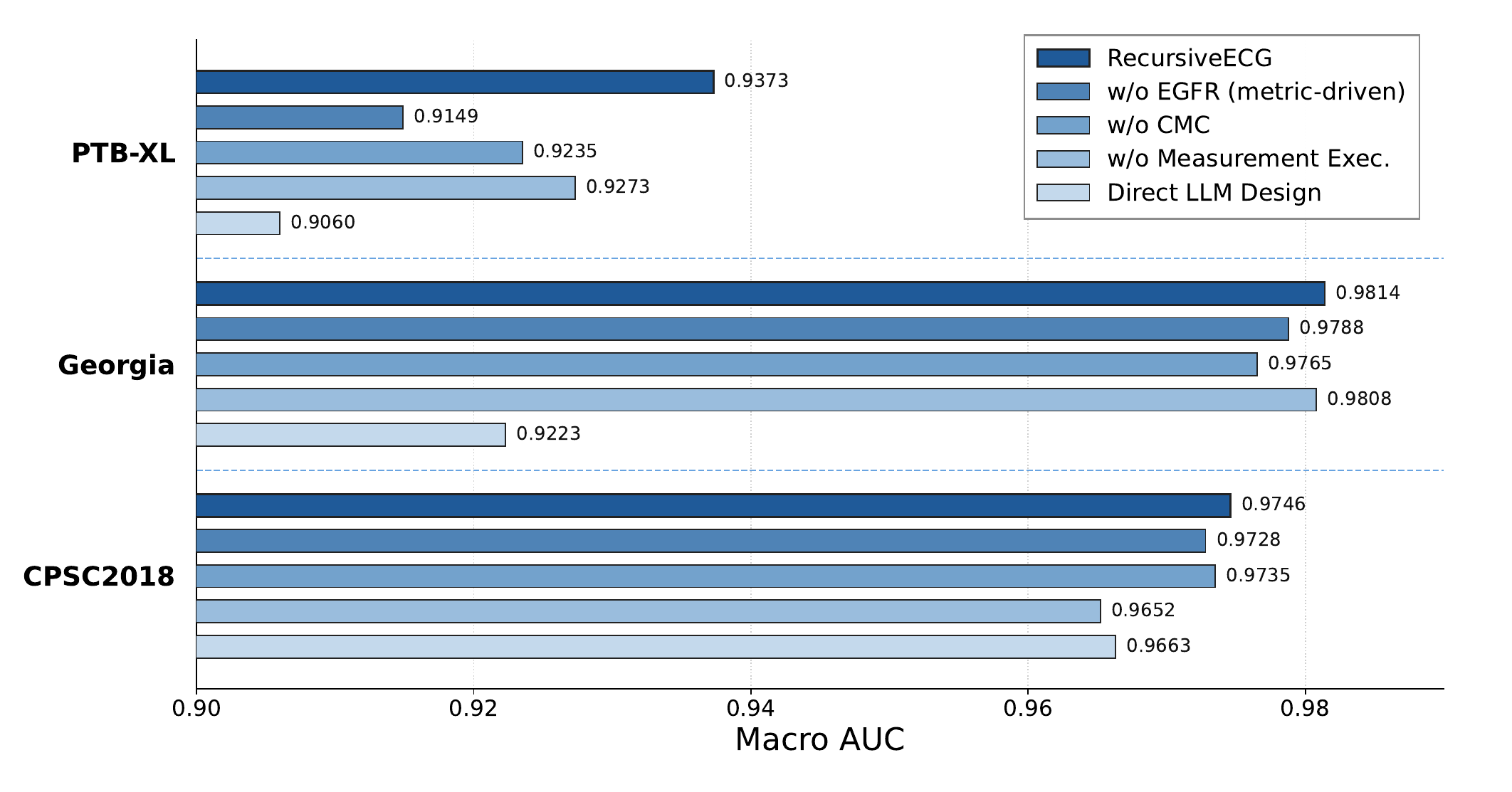}
\Description{Grouped bar chart showing absolute Macro AUC for \sysname{} and five ablation variants: w/o EGFR, w/o CMC, w/o Measurement Execution, Direct LLM Design, across PTB-XL, Georgia, and CPSC2018.}
\caption{Mechanism ablation study across three datasets.}
\label{fig:main_ablation_macro_drop}
\end{figure}

\begin{figure*}[htbp]
\centering
\includegraphics[width=\textwidth]{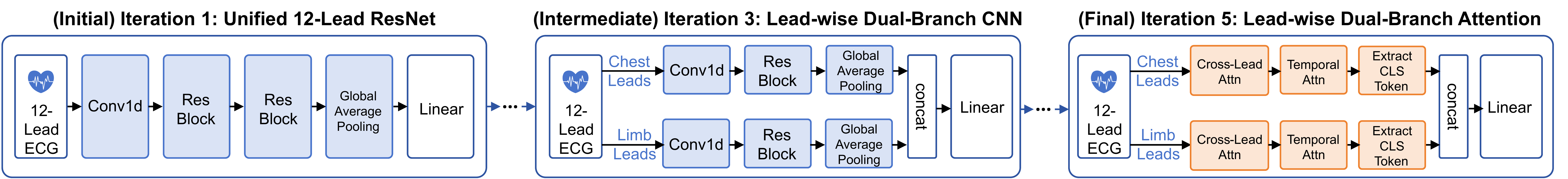}\hfill
\Description{Evolution of the ECG classification architecture from a unified 12-lead ResNet to a dual-branch CNN that separately processes precordial and limb leads, followed by a lead-wise dual-branch attention model that learns lead-specific weights and inter-lead dependencies before feature fusion and classification.}
\caption{Model design evolution on PTB-XL.}
\label{fig:model_design_evolution_ptbxl}
\end{figure*}

\begin{figure}[t]
\centering
\includegraphics[width=\linewidth]{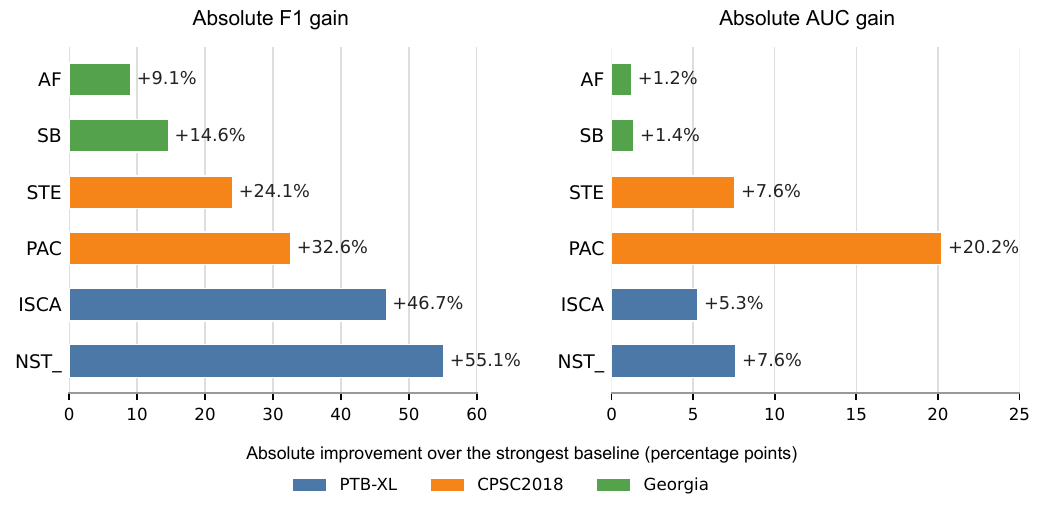}
\caption{Representative gains on fine-grained labels. Values indicate the absolute gains of \sysname{} over the strongest baseline for Macro F1 and AUC in percentage.}
\label{fig:fine_grained_label_gains}
\end{figure}

\subsection{Main Results}

We evaluate the classifiers discovered by \sysname{} from two perspectives: predictive effectiveness and computational efficiency.

\fakeparagraph{Overall Performance Improvement}
Table~\ref{tab:main_aggregate} summarizes the performance comparison on PTB-XL, Georgia, and CPSC2018 under two backbone LLMs.
\sysname{} consistently achieves the best performance across all datasets and evaluation metrics compared with other non-\sysname{} baselines, outperforming representative deep models, pretrained models, and agentic approaches.
For Macro F1, \sysname{} improves by 5.56\%, 8.56\%, and 15.81\% over the strongest baselines on PTB-XL, Georgia, and CPSC2018, respectively.
Moreover, \sysname exhibits well-controlled variance across independent runs, demonstrating that it can reliably evolve high-performing architectures.
Using Qwen3.5-27B \cite{qwen3.5} as an alternative LLM backbone also achieves competitive performance, indicating that the proposed evidence-grounded refinement paradigm remains effective across different backbone LLMs.

\fakeparagraph{Effectiveness--Efficiency Trade-off}
We evaluate whether the classifiers discovered by \sysname achieve a favorable effectiveness--efficiency trade-off. Figure~\ref{fig:ptbxl_efficiency_bubble} compares representative classifiers in terms of predictive performance and computational cost.
\sysname achieves superior predictive effectiveness with 10.6\% fewer parameters than xLSTM-ECG and 90.5\% fewer FLOPs than UniTS, while reducing peak memory by 65.8\% relative to MERL.
These advantages indicate that its performance gains do not arise from increased model complexity.

\subsection{Ablation Study}
\label{sec:ablation}

To determine whether the gains of \sysname{} arise from LLM-based model design or from its refinement mechanisms, we conduct ablation studies to assess the roles of all components in \sysname{}.
We evaluate variants without Evidence-Grounded Failure Review (w/o EGFR), Criteria-to-Measurement Compilation (w/o CMC), or measurement execution (w/o Measurement Exec.), together with a Direct LLM Design variant that removes evidence-grounded failure review and iterative refinement.
Detailed definitions are provided in \secref{app:ablation_latest}.

As shown in \figref{fig:main_ablation_macro_drop}, the complete framework provides the best performance, although the contribution of each mechanism varies across datasets.
Interestingly, Direct LLM Design remains competitive and even outperforms several existing baselines, demonstrating the capability of LLMs for initial model design. However, its gap with \sysname{} shows that LLM knowledge alone is insufficient without evidence-grounded diagnosis and iterative refinement.

Removing EGFR causes notable degradation on PTB-XL, indicating that metric-driven refinement cannot provide sufficient guidance for identifying model limitations and generating targeted revisions.
Disabling measurement execution causes a measurable but relatively small drop on CPSC2018, while its effect varies across datasets.
CMC further provides complementary benefits by translating clinical criteria into structured measurements.
Their effects are complementary rather than interchangeable within the iterative evolution loop.

These results show that effective ECG classifier evolution depends not only on LLM-based design or metric optimization, but on a structured, evidence-grounded process of failure diagnosis, executable validation, and iterative refinement across datasets.

\begin{figure}[t]
\centering
\includegraphics[width=\linewidth]{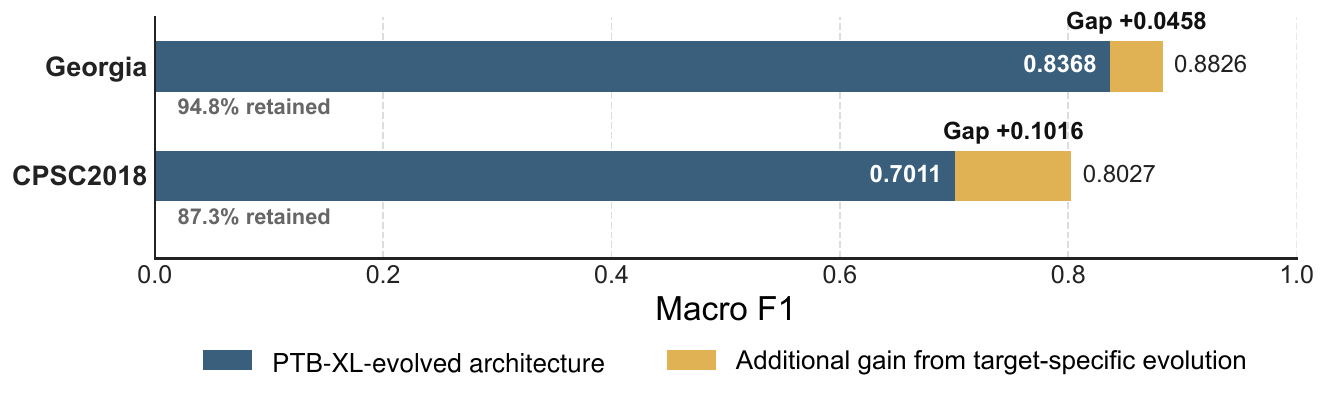}
\caption{Dataset-adaptive architecture evolution.}
\label{fig:arch_transfer}
\end{figure}

\subsection{Model Design Evolution}
\label{sec:model_design_evolution_ptbxl}

To understand whether \sysname{} discovers meaningful model designs, we analyze the evolution trajectory of the ECG classifier on PTB-XL.
As shown in \figref{fig:model_design_evolution_ptbxl}, \sysname{} progressively evolves a generic multi-lead architecture into an ECG-specific design by introducing lead-aware representation learning and dependency modeling across leads and temporal patterns.
Starting from a unified 12-lead ResNet, \sysname{} first separates chest and limb leads to capture heterogeneous lead information, and then incorporates cross-lead and temporal attention to enhance feature interaction.
This evolution trajectory demonstrates that \sysname{} performs targeted architecture refinement rather than arbitrary model modifications, progressively discovering ECG-specific inductive biases through evidence-driven evolution.



\subsection{Fine-Grained Label Behavior}

\begin{figure*}[htbp]
\centering
\includegraphics[width=\textwidth]{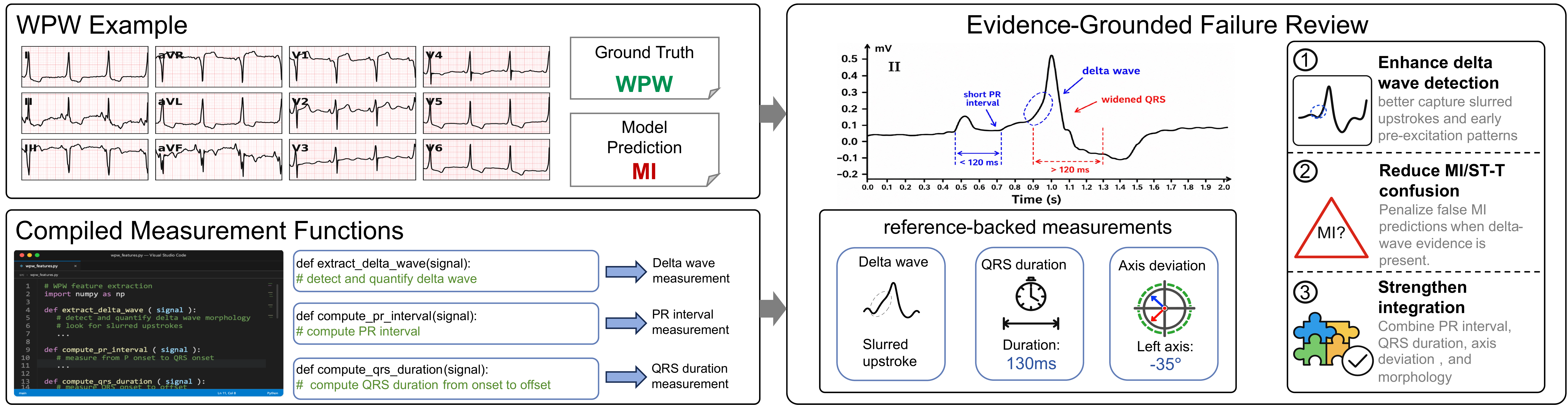}\hfill
\Description{WPW case visualization showing pseudo-infarction morphology, reference-backed measurements, cross-lead evidence, and the resulting evidence-supported revision guidance.}
\caption{Evidence-Grounded Failure Review of a WPW pseudo-infarction case.}
\label{fig:case_wpw_pseudo_mi}
\end{figure*}



While aggregate metrics demonstrate the effectiveness of \sysname, they do not show how gains are distributed across diagnostic categories. We therefore conduct a fine-grained label-level analysis. \figref{fig:fine_grained_label_gains} highlights representative categories with notable improvements, while complete label-wise and negative-label results are provided in \secref{app:fine_grained_latest} and \secref{app:failure_modes_latest}.

The gains are most pronounced in diagnoses with explicit, clinically interpretable ECG patterns. On PTB-XL, \sysname improves substantially on NST\_ and ISCA, which depend on local waveform morphology and ST-T characteristics. It also achieves notable gains on PAC in CPSC2018 and on rhythm-based diagnoses such as SB and AF in Georgia. These results suggest that diagnoses with well-defined ECG characteristics are particularly suitable for evidence-grounded model evolution, as their failure modes can be described using clinical measurements and case-level evidence.

\subsection{Dataset-Adaptive Architecture Evolution}

An important capability of an autonomous model designer is to adapt its design strategy according to the characteristics of different datasets.
To investigate this capability, we perform dataset-adaptive architecture evolution experiments.
Specifically, we first evolve an ECG classifier on PTB-XL and apply the discovered architecture to other datasets, where model parameters are retrained using the corresponding target dataset.
We then further evolve the architecture with \sysname on each target dataset and evaluate the additional improvements obtained from dataset-specific evolution.

As shown in \figref{fig:arch_transfer}, the architecture evolved on PTB-XL already provides a strong architectural prior for other datasets, achieving Macro F1 scores of 0.8368 and 0.7011 on Georgia and CPSC2018, respectively; the complete transfer results are reported in \tabref{tab:app_arch_transfer_latest}.
After target-specific evolution, \sysname{} further improves Macro F1 by 0.0458 and 0.1016 on Georgia and CPSC2018, reaching 0.8826 and 0.8027.

These results demonstrate that \sysname{} discovers reusable ECG modeling patterns while adapting designs to dataset-specific characteristics, rather than relying on a fixed architecture.

\subsection{Case Study}
\label{sec:case_review_model_limitation_ptbxl}

We use a WPW pseudo-infarction case from PTB-XL to show how evidence-grounded review converts a failed prediction into an identified model limitation.
The case analysis is shown in \figref{fig:case_wpw_pseudo_mi}, with the full audit trace in \tabref{tab:app_wpw_case_review}.

The initial unified 12-lead ResNet jointly encoded all leads but did not explicitly distinguish their spatial and diagnostic roles.
In this case, WPW-related pre-excitation produced infarction-like initial deflections, causing the model to overemphasize local morphology and incorrectly predict myocardial infarction.

The review identified a short PR interval, slurred QRS onset, and QRS widening, all consistent with WPW.
Moreover, the apparent infarction pattern conflicted with the overall electrical axis and ventricular activation pattern.
The failure therefore revealed a model limitation: insufficient use of complementary cross-lead evidence to reconcile local morphology with global activation patterns.

\balance

\section{Conclusion}
\label{sec:conclusion}

Metric-driven feedback can evaluate whether a model improves, but provides limited guidance for understanding failures and refining model designs. 
In this work, we introduce \sysname, an evidence-driven LLM-as-Designer paradigm that transforms prediction failures into actionable evidence and executable revisions for recursive ECG classifier refinement. Experiments on three ECG benchmarks demonstrate that RecursiveECG achieves consistent performance improvements, efficient deployment, and dataset-adaptive evolution by discovering ECG-specific inductive biases.

Future work will explore extending this paradigm toward a general-purpose medical time-series model refinement agent, capable of leveraging task-specific failures and domain evidence to autonomously improve predictive models across diverse physiological signals and healthcare applications.


\section{Limitations and Ethical Considerations}
Our experiments use publicly released, de-identified ECG benchmarks under their respective licenses. 
\sysname may inherit errors or biases from the underlying LLM and curated clinical criteria. 
The system has only been evaluated retrospectively and is not intended for autonomous diagnosis or clinical deployment without external clinician oversight.

\section{Generative AI Usage}
DeepSeek-V4-Pro served as the default backbone LLM for the offline \sysname{} workflow. 
Qwen3.5-27B was used in the LLM-backbone sensitivity experiments, and ChatGPT was used to assist with language polishing and graphical icon editing in the manuscript.
The authors verified all outputs and accept responsibility.

\clearpage
\bibliographystyle{ACM-Reference-Format}
\bibliography{ref}

@article{gu2025argos,
  title={{Argos}: Agentic Time-Series Anomaly Detection with Autonomous Rule Generation via Large Language Models},
  author={Gu, Yile and Xiong, Yifan and Mace, Jonathan and Jiang, Yuting and Hu, Yigong and Kasikci, Baris and Cheng, Peng},
  journal={arXiv preprint arXiv:2501.14170},
  year={2025}
}

@article{yue2022ts2vec,
  title={{TS2Vec}: Towards Universal Representation of Time Series},
  author={Yue, Zhihan and Wang, Yujing and Duan, Juanyong and Yang, Tianmeng and Huang, Congrui and Tong, Yunhai and Xu, Bixiong},
  journal={Proceedings of the AAAI Conference on Artificial Intelligence},
  volume={36},
  number={8},
  pages={8980--8987},
  year={2022},
  doi={10.1609/aaai.v36i8.20881}
}

@inproceedings{wu2023timesnet,
  title={{TimesNet}: Temporal 2D-Variation Modeling for General Time Series Analysis},
  author={Wu, Haixu and Hu, Tengge and Liu, Yong and Zhou, Hang and Wang, Jianmin and Long, Mingsheng},
  booktitle={The Eleventh International Conference on Learning Representations},
  year={2023}
}

@inproceedings{nie2023patchtst,
  title={A Time Series is Worth 64 Words: Long-term Forecasting with Transformers},
  author={Nie, Yuqi and Nguyen, Nam H. and Sinthong, Phanwadee and Kalagnanam, Jayant},
  booktitle={The Eleventh International Conference on Learning Representations},
  year={2023}
}

@inproceedings{gao2024units,
  title={{UniTS}: A Unified Multi-Task Time Series Model},
  author={Gao, Shanghua and Koker, Teddy and Queen, Owen and Hartvigsen, Thomas and Tsiligkaridis, Theodoros and Zitnik, Marinka},
  booktitle={Advances in Neural Information Processing Systems},
  volume={37},
  year={2024}
}

@article{alday2020classification,
  title={Classification of 12-lead ecgs: the physionet/computing in cardiology challenge 2020},
  author={{Perez Alday}, Erick A. and Gu, Annie and Shah, Amit J. and Robichaux, Chad and Wong, An-Kwok Ian and Liu, Chengyu and Liu, Feifei and {Bahrami Rad}, Ali and Elola, Andoni and Seyedi, Salman and Li, Qiao and Sharma, Ashish and Clifford, Gari D. and Reyna, Matthew A.},
  volume={41},
  number={12},
  pages={124003},
  year={2020},
  publisher={IOP Publishing}
}

@article{wagner2020ptbxl,
  title={PTB-XL, a large publicly available electrocardiography dataset},
  author={Wagner, Patrick and Strodthoff, Nils and Bousseljot, Ralf-Dieter and Kreiseler, Dieter and Lunze, Fatima I and Samek, Wojciech and Schaeffter, Tobias},
  journal={Scientific data},
  volume={7},
  number={1},
  pages={154},
  year={2020},
  publisher={Nature Publishing Group UK London}
}

@article{strodthoff2021deep,
  title={Deep Learning for ECG Analysis: Benchmarks and Insights from {PTB-XL}},
  author={Strodthoff, Nils and Wagner, Patrick and Schaeffter, Tobias and Samek, Wojciech},
  journal={IEEE Journal of Biomedical and Health Informatics},
  volume={25},
  number={5},
  pages={1519--1528},
  year={2021},
  publisher={IEEE},
  doi={10.1109/JBHI.2020.3022989},
  url={https://doi.org/10.1109/JBHI.2020.3022989}
}

@inproceedings{na2024stmem,
  title={Guiding Masked Representation Learning to Capture Spatio-Temporal Relationship of Electrocardiogram},
  author={Na, Yeongyeon and Park, Minje and Tae, Yunwon and Joo, Sunghoon},
  booktitle={The Twelfth International Conference on Learning Representations},
  year={2024}
}

@inproceedings{liu2024merl,
  title={Zero-Shot ECG Classification with Multimodal Learning and Test-time Clinical Knowledge Enhancement},
  author={Liu, Che and Wan, Zhongwei and Ouyang, Cheng and Shah, Anand and Bai, Wenjia and Arcucci, Rossella},
  booktitle={International Conference on Machine Learning},
  pages={31949--31963},
  year={2024},
  organization={PMLR}
}

@inproceedings{zhou2025htuning,
  title={H-Tuning: Toward Low-Cost and Efficient ECG-based Cardiovascular Disease Detection with Pre-Trained Models},
  author={Zhou, Rushuang and Zhang, Yuanting and Dong, Yining},
  booktitle={International Conference on Machine Learning},
  pages={79548--79569},
  year={2025},
  organization={PMLR}
}

@inproceedings{wang2025melp,
  title={From Token to Rhythm: A Multi-Scale Approach for ECG-Language Pretraining},
  author={Wang, Fuying and Xu, Jiacheng and Yu, Lequan},
  booktitle={International Conference on Machine Learning},
  pages={65059--65074},
  year={2025},
  organization={PMLR}
}

@article{kang2025xlstmecg,
  title={{xLSTM-ECG}: Multi-label {ECG} Classification via Feature Fusion with {xLSTM}},
  author={Kang, Lei and Fu, Xuanshuo and Vazquez-Corral, Javier and Valveny, Ernest and Karatzas, Dimosthenis},
  journal={arXiv preprint arXiv:2504.16101},
  year={2025}
}

@inproceedings{oh2023ecgqa,
  title={{ECG-QA}: A Comprehensive Question Answering Dataset Combined With Electrocardiogram},
  author={Oh, Jungwoo and Lee, Gyubok and Bae, Seongsu and Kwon, Joon-myoung and Choi, Edward},
  booktitle={Advances in Neural Information Processing Systems, Datasets and Benchmarks Track},
  volume={36},
  pages={66277--66288},
  year={2023}
}

@inproceedings{wan2025meit,
  title={MEIT: Multimodal electrocardiogram instruction tuning on large language models for report generation},
  author={Wan, Zhongwei and Liu, Che and Wang, Xin and Tao, Chaofan and Shen, Hui and Xiong, Jing and Arcucci, Rossella and Yao, Huaxiu and Zhang, Mi},
  booktitle={Findings of the association for computational linguistics: ACL 2025},
  pages={14510--14527},
  publisher={Association for Computational Linguistics},
  address={Vienna, Austria},
  doi={10.18653/v1/2025.findings-acl.749},
  year={2025}
}

@inproceedings{xu2026visualtime,
  title={Can Multimodal {LLM}s Perform Time Series Anomaly Detection?},
  author={Xu, Xiongxiao and Wang, Haoran and Liang, Yueqing and Yu, Philip S. and Zhao, Yue and Shu, Kai},
  booktitle={Proceedings of the ACM Web Conference 2026},
  pages={5392--5403},
  year={2026},
  doi={10.1145/3774904.3792376},
  url={https://doi.org/10.1145/3774904.3792376}
}

@inproceedings{park2025delving,
  title={Delving into Large Language Models for Effective Time-Series Anomaly Detection},
  author={Park, Junwoo and Jung, Kyudan and Lee, Dohyun and Lee, Hyuck and Gwak, Daehoon and Park, ChaeHun and Choo, Jaegul and Cho, Jaewoong},
  booktitle={Advances in Neural Information Processing Systems},
  volume={38},
  year={2025},
  url={https://proceedings.neurips.cc/paper_files/paper/2025/hash/950a34c7b107111570077078e1b5b8ef-Abstract-Conference.html}
}

@inproceedings{yang2025adagent,
  title={AD-AGENT: A Multi-agent Framework for End-to-end Anomaly Detection},
  author={Yang, Tiankai and Liu, Junjun and Siu, Michael and Wang, Jiahang and Qian, Zhuangzhuang and Song, Chanjuan and Cheng, Cheng and Hu, Xiyang and Zhao, Yue},
  booktitle={Proceedings of the 14th International Joint Conference on Natural Language Processing and the 4th Conference of the Asia-Pacific Chapter of the Association for Computational Linguistics},
  pages={191--205},
  year={2025},
  publisher={The Asian Federation of Natural Language Processing and
           The Association for Computational Linguistics},
  address={Mumbai, India},
  doi={10.18653/v1/2025.findings-ijcnlp.11}
}

@article{tao2026anomamind,
  title={{AnomaMind}: Agentic Time Series Anomaly Detection with Tool-Augmented Reasoning},
  author={Tao, Xiaoyu and Wu, Yuchong and Cheng, Mingyue and Guo, Ze and Gao, Tian},
  journal={arXiv preprint arXiv:2602.13807},
  year={2026},
  doi={10.48550/arXiv.2602.13807},
  url={https://arxiv.org/abs/2602.13807}
}

@article{gruver2023large,
  title={Large language models are zero-shot time series forecasters},
  author={Gruver, Nate and Finzi, Marc and Qiu, Shikai and Wilson, Andrew G},
  journal={Advances in neural information processing systems},
  volume={36},
  pages={19622--19635},
  year={2023}
}

@inproceedings{yao2023react,
  title     = {{ReAct}: Synergizing Reasoning and Acting in Language Models},
  author    = {Yao, Shunyu and Zhao, Jeffrey and Yu, Dian and Du, Nan and Shafran, Izhak and Narasimhan, Karthik and Cao, Yuan},
  booktitle = {The Eleventh International Conference on Learning Representations},
  year      = {2023},
  url       = {https://openreview.net/forum?id=WE_vluYUL-X}
}

@inproceedings{shinn2023reflexion,
  title     = {Reflexion: Language Agents with Verbal Reinforcement Learning},
  author    = {Shinn, Noah and Cassano, Federico and Gopinath, Ashwin and Narasimhan, Karthik and Yao, Shunyu},
  booktitle = {Advances in Neural Information Processing Systems},
  volume    = {36},
  year      = {2023}
}

@article{schick2023toolformer,
  title={Toolformer: Language models can teach themselves to use tools},
  author={Schick, Timo and Dwivedi-Yu, Jane and Dess{\`\i}, Roberto and Raileanu, Roberta and Lomeli, Maria and Hambro, Eric and Zettlemoyer, Luke and Cancedda, Nicola and Scialom, Thomas},
  journal={Advances in neural information processing systems},
  volume={36},
  pages={68539--68551},
  year={2023}
}

@article{kligfield2007ecgpart1,
  author   = {Kligfield, Paul and Gettes, Leonard S. and Bailey, James J. and
              Childers, Rory and Deal, Barbara J. and Hancock, E. William and
              van Herpen, Gerard and Kors, Jan A. and Macfarlane, Peter and
              Mirvis, David M. and Pahlm, Olle and Rautaharju, Pentti and
              Wagner, Galen S. and Josephson, Mark and Mason, Jay W. and
              Okin, Peter and Surawicz, Borys and Wellens, Hein and
              {{American Heart Association Electrocardiography and Arrhythmias Committee,
              Council on Clinical Cardiology}} and
              {{American College of Cardiology Foundation}} and
              {{Heart Rhythm Society}}},
  title    = {Recommendations for the Standardization and Interpretation of the
              Electrocardiogram: Part I: The Electrocardiogram and Its Technology:
              A Scientific Statement from the American Heart Association
              Electrocardiography and Arrhythmias Committee, Council on Clinical
              Cardiology; the American College of Cardiology Foundation; and the
              Heart Rhythm Society Endorsed by the International Society for
              Computerized Electrocardiology},
  journal  = {Journal of the American College of Cardiology},
  year     = {2007},
  month    = mar,
  volume   = {49},
  number   = {10},
  pages    = {1109--1127},
  doi      = {10.1016/j.jacc.2007.01.024},
  issn     = {1558-3597},
  pmid     = {17349896},
  language = {English}
}

@article{mason2007ecgpart2,
  title={{Recommendations for the Standardization and Interpretation of the Electrocardiogram: Part II: Electrocardiography Diagnostic Statement List: A Scientific Statement From the American Heart Association Electrocardiography and Arrhythmias Committee, Council on Clinical Cardiology; the American College of Cardiology Foundation; and the Heart Rhythm Society: Endorsed by the International Society for Computerized Electrocardiology}},
  author={Mason, Jay W. and Hancock, E. William and Gettes, Leonard S. and Bailey, James J. and Childers, Rory and Deal, Barbara J. and Josephson, Mark and Kligfield, Paul and Kors, Jan A. and Macfarlane, Peter and Pahlm, Olle and Mirvis, David M. and Okin, Peter and Rautaharju, Pentti M. and Surawicz, Borys and van Herpen, Gerard and Wagner, Galen S. and Wellens, Hein},
  journal={Journal of the American College of Cardiology},
  volume={49},
  number={10},
  pages={1128--1135},
  year={2007},
  doi={10.1016/j.jacc.2007.01.025}
}

@article{surawicz2009ecgpart3,
  title={{{AHA/ACCF/HRS} Recommendations for the Standardization and Interpretation of the Electrocardiogram: Part III: Intraventricular Conduction Disturbances: A Scientific Statement From the American Heart Association Electrocardiography and Arrhythmias Committee, Council on Clinical Cardiology; the American College of Cardiology Foundation; and the Heart Rhythm Society: Endorsed by the International Society for Computerized Electrocardiology}},
  author={Surawicz, Borys and Childers, Rory and Deal, Barbara J. and Gettes, Leonard S. and Bailey, James J. and Gorgels, Anton and Hancock, E. William and Josephson, Mark and Kligfield, Paul and Kors, Jan A. and Macfarlane, Peter and Mason, Jay W. and Mirvis, David M. and Okin, Peter and Pahlm, Olle and Rautaharju, Pentti M. and van Herpen, Gerard and Wagner, Galen S. and Wellens, Hein},
  journal={Journal of the American College of Cardiology},
  volume={53},
  number={11},
  pages={976--981},
  year={2009},
  doi={10.1016/j.jacc.2008.12.013}
}

@article{rautaharju2009ecgpart4,
  title={{{AHA/ACCF/HRS} Recommendations for the Standardization and Interpretation of the Electrocardiogram: Part IV: The ST Segment, T and U Waves, and the QT Interval: A Scientific Statement From the American Heart Association Electrocardiography and Arrhythmias Committee, Council on Clinical Cardiology; the American College of Cardiology Foundation; and the Heart Rhythm Society: Endorsed by the International Society for Computerized Electrocardiology}},
  author={Rautaharju, Pentti M. and Surawicz, Borys and Gettes, Leonard S. and Bailey, James J. and Childers, Rory and Deal, Barbara J. and Gorgels, Anton and Hancock, E. William and Josephson, Mark and Kligfield, Paul and Kors, Jan A. and Macfarlane, Peter and Mason, Jay W. and Mirvis, David M. and Okin, Peter and Pahlm, Olle and van Herpen, Gerard and Wagner, Galen S. and Wellens, Hein},
  journal={Journal of the American College of Cardiology},
  volume={53},
  number={11},
  pages={982--991},
  year={2009},
  doi={10.1016/j.jacc.2008.12.014}
}

@article{hancock2009ecgpart5,
  title={{{AHA/ACCF/HRS} Recommendations for the Standardization and Interpretation of the Electrocardiogram: Part V: Electrocardiogram Changes Associated With Cardiac Chamber Hypertrophy: A Scientific Statement From the American Heart Association Electrocardiography and Arrhythmias Committee, Council on Clinical Cardiology; the American College of Cardiology Foundation; and the Heart Rhythm Society: Endorsed by the International Society for Computerized Electrocardiology}},
  author={Hancock, E. William and Deal, Barbara J. and Mirvis, David M. and Okin, Peter and Kligfield, Paul and Gettes, Leonard S. and Bailey, James J. and Childers, Rory and Gorgels, Anton and Josephson, Mark and Kors, Jan A. and Macfarlane, Peter and Mason, Jay W. and Pahlm, Olle and Rautaharju, Pentti M. and Surawicz, Borys and van Herpen, Gerard and Wagner, Galen S. and Wellens, Hein},
  journal={Journal of the American College of Cardiology},
  volume={53},
  number={11},
  pages={992--1002},
  year={2009},
  doi={10.1016/j.jacc.2008.12.015}
}

@article{wagner2009ecgpart6,
  title={{{AHA/ACCF/HRS} Recommendations for the Standardization and Interpretation of the Electrocardiogram: Part VI: Acute Ischemia/Infarction: A Scientific Statement From the American Heart Association Electrocardiography and Arrhythmias Committee, Council on Clinical Cardiology; the American College of Cardiology Foundation; and the Heart Rhythm Society: Endorsed by the International Society for Computerized Electrocardiology}},
  author={Wagner, Galen S. and Macfarlane, Peter and Wellens, Hein and Josephson, Mark and Gorgels, Anton and Mirvis, David M. and Pahlm, Olle and Surawicz, Borys and Kligfield, Paul and Childers, Rory and Gettes, Leonard S. and Bailey, James J. and Deal, Barbara J. and Hancock, E. William and Kors, Jan A. and Mason, Jay W. and Okin, Peter and Rautaharju, Pentti M. and van Herpen, Gerard},
  journal={Journal of the American College of Cardiology},
  volume={53},
  number={11},
  pages={1003--1011},
  year={2009},
  doi={10.1016/j.jacc.2008.12.016}
}

@inproceedings{huang2024mlagentbench,
  title     = {{MLAgentBench}: Evaluating Language Agents on Machine Learning Experimentation},
  author    = {Huang, Qian and Vora, Jian and Liang, Percy and Leskovec, Jure},
  booktitle = {Proceedings of the 41st International Conference on Machine Learning},
  pages     = {20271--20309},
  year      = {2024},
  volume    = {235},
  series    = {Proceedings of Machine Learning Research},
  publisher = {PMLR},
  url       = {https://proceedings.mlr.press/v235/huang24y.html}
}

@inproceedings{ma2024llmsimulation,
  title     = {{LLM} and Simulation as Bilevel Optimizers: A New Paradigm to Advance Physical Scientific Discovery},
  author    = {Ma, Pingchuan and Wang, Tsun-Hsuan and Guo, Minghao and Sun, Zhiqing and Tenenbaum, Joshua B. and Rus, Daniela and Gan, Chuang and Matusik, Wojciech},
  booktitle = {Proceedings of the 41st International Conference on Machine Learning},
  pages     = {33940--33962},
  year      = {2024},
  volume    = {235},
  series    = {Proceedings of Machine Learning Research},
  publisher = {PMLR},
  url       = {https://proceedings.mlr.press/v235/ma24m.html}
}

@inproceedings{shojaee2025llmsr,
  title     = {{LLM-SR}: Scientific Equation Discovery via Programming with Large Language Models},
  author    = {Shojaee, Parshin and Meidani, Kazem and Gupta, Shashank and Barati Farimani, Amir and Reddy, Chandan K.},
  booktitle = {The Thirteenth International Conference on Learning Representations},
  year      = {2025},
  url       = {https://proceedings.iclr.cc/paper_files/paper/2025/hash/28df8e730c054c5331855fd4d5403ba9-Abstract-Conference.html}
}

@article{novikov2025alphaevolve,
  title   = {{AlphaEvolve}: A Coding Agent for Scientific and Algorithmic Discovery},
  author  = {Novikov, Alexander and V{\~u}, Ng{\^a}n and Eisenberger, Marvin and Dupont, Emilien and Huang, Po-Sen and Wagner, Adam Zsolt and Shirobokov, Sergey and Kozlovskii, Borislav and Ruiz, Francisco J. R. and Mehrabian, Abbas and Kumar, M. Pawan and See, Abigail and Chaudhuri, Swarat and Holland, George and Davies, Alex and Nowozin, Sebastian and Kohli, Pushmeet and Balog, Matej},
  journal = {arXiv preprint arXiv:2506.13131},
  year    = {2025},
  doi     = {10.48550/arXiv.2506.13131},
  url     = {https://arxiv.org/abs/2506.13131}
}

@inproceedings{huang2025popper,
  title     = {Automated Hypothesis Validation with Agentic Sequential Falsifications},
  author    = {Huang, Kexin and Jin, Ying and Li, Ryan and Li, Michael Y. and Candes, Emmanuel and Leskovec, Jure},
  booktitle = {Proceedings of the 42nd International Conference on Machine Learning},
  pages     = {25372--25437},
  year      = {2025},
  volume    = {267},
  series    = {Proceedings of Machine Learning Research},
  publisher = {PMLR},
  url       = {https://proceedings.mlr.press/v267/huang25n.html}
}

@inproceedings{jansen2024discoveryworld,
  title     = {{DiscoveryWorld}: A Virtual Environment for Developing and Evaluating Automated Scientific Discovery Agents},
  author    = {Jansen, Peter A. and C{\^o}t{\'e}, Marc-Alexandre and Khot, Tushar and Bransom, Erin and Mishra, Bhavana Dalvi and Majumder, Bodhisattwa Prasad and Tafjord, Oyvind and Clark, Peter},
  booktitle = {Advances in Neural Information Processing Systems},
  pages     = {10088--10116},
  year      = {2024},
  volume    = {37},
  publisher = {Curran Associates, Inc.},
  doi       = {10.52202/079017-0324},
  url       = {https://proceedings.neurips.cc/paper_files/paper/2024/hash/13836f251823945316ae067350a5c366-Abstract-Datasets_and_Benchmarks_Track.html}
}

@misc{deepseekai2026deepseekv4,
      title={DeepSeek-V4: Towards Highly Efficient Million-Token Context Intelligence},
      author={DeepSeek-AI},
      year={2026},
}

@misc{qwen3.5,
    title  = {{Qwen3.5}: Towards Native Multimodal Agents},
    author = {{Qwen Team}},
    month  = {February},
    year   = {2026},
    url    = {https://qwen.ai/blog?id=qwen3.5}
}

@article{hannun2019cardiologist,
  title   = {Cardiologist-Level Arrhythmia Detection and Classification in Ambulatory Electrocardiograms Using a Deep Neural Network},
  author  = {Hannun, Awni Y. and Rajpurkar, Pranav and Haghpanahi, Masoumeh and Tison, Geoffrey H. and Bourn, Codie and Turakhia, Mintu P. and Ng, Andrew Y.},
  journal = {Nature Medicine},
  volume  = {25},
  number  = {1},
  pages   = {65--69},
  year    = {2019},
  doi     = {10.1038/s41591-018-0268-3}
}

@article{attia2019screening,
  title   = {Screening for Cardiac Contractile Dysfunction Using an Artificial Intelligence-Enabled Electrocardiogram},
  author  = {Attia, Zachi I. and Kapa, Suraj and Lopez-Jimenez, Francisco and McKie, Paul M. and Ladewig, Dorothy J. and Satam, Gaurav and Pellikka, Patricia A. and Enriquez-Sarano, Maurice and Noseworthy, Peter A. and Munger, Thomas M. and Asirvatham, Samuel J. and Scott, Christopher G. and Carter, Rickey E. and Friedman, Paul A.},
  journal = {Nature Medicine},
  volume  = {25},
  number  = {1},
  pages   = {70--74},
  year    = {2019},
  doi     = {10.1038/s41591-018-0240-2}
}

@article{ribeiro2020automatic,
  title   = {Automatic Diagnosis of the 12-Lead {ECG} Using a Deep Neural Network},
  author  = {Ribeiro, Ant{\^o}nio H. and Ribeiro, Manoel Horta and Paix{\~a}o, Gabriela M. M. and Oliveira, Derick M. and Gomes, Paulo R. and Canazart, J{\'e}ssica A. and Ferreira, Milton P. S. and Andersson, Carl R. and Macfarlane, Peter W. and Meira Jr., Wagner and Sch{\"o}n, Thomas B. and Ribeiro, Antonio Luiz P.},
  journal = {Nature Communications},
  volume  = {11},
  pages   = {1760},
  year    = {2020},
  doi     = {10.1038/s41467-020-15432-4}
}

@article{zhu2020automatic,
  title   = {Automatic Multilabel Electrocardiogram Diagnosis of Heart Rhythm or Conduction Abnormalities with Deep Learning: A Cohort Study},
  author  = {Zhu, Hongling and Cheng, Cheng and Yin, Hang and Li, Xingyi and Zuo, Ping and Ding, Jia and Lin, Fan and Wang, Jingyi and Zhou, Beitong and Li, Yonge and Hu, Shouxing and Xiong, Yulong and Wang, Binran and Wan, Guohua and Yang, Xiaoyun and Yuan, Ye},
  journal = {The Lancet Digital Health},
  volume  = {2},
  number  = {7},
  pages   = {e348--e357},
  year    = {2020},
  doi     = {10.1016/S2589-7500(20)30107-2}
}

@inproceedings{kiyasseh2021clocs,
  title     = {{CLOCS}: Contrastive Learning of Cardiac Signals Across Space, Time, and Patients},
  author    = {Kiyasseh, Dani and Zhu, Tingting and Clifton, David A.},
  booktitle = {Proceedings of the 38th International Conference on Machine Learning},
  series    = {Proceedings of Machine Learning Research},
  volume    = {139},
  pages     = {5606--5615},
  year      = {2021},
  publisher = {PMLR},
  url       = {https://proceedings.mlr.press/v139/kiyasseh21a.html}
}

@inproceedings{golany2019pgans,
  title     = {{PGANs}: Personalized Generative Adversarial Networks for {ECG} Synthesis to Improve Patient-Specific Deep {ECG} Classification},
  author    = {Golany, Tomer and Radinsky, Kira},
  booktitle = {Proceedings of the AAAI Conference on Artificial Intelligence},
  volume    = {33},
  number    = {1},
  pages     = {557--564},
  year      = {2019},
  doi       = {10.1609/aaai.v33i01.3301557}
}

@inproceedings{golany2020improving,
  title     = {Improving {ECG} Classification Using Generative Adversarial Networks},
  author    = {Golany, Tomer and Lavee, Gal and Tejman Yarden, Shai and Radinsky, Kira},
  booktitle = {Proceedings of the AAAI Conference on Artificial Intelligence},
  volume    = {34},
  number    = {8},
  pages     = {13280--13285},
  year      = {2020},
  doi       = {10.1609/aaai.v34i08.7037}
}

@inproceedings{jin2024timellm,
  title     = {{Time-LLM}: Time Series Forecasting by Reprogramming Large Language Models},
  author    = {Jin, Ming and Wang, Shiyu and Ma, Lintao and Chu, Zhixuan and Zhang, James Y. and Shi, Xiaoming and Chen, Pin-Yu and Liang, Yuxuan and Li, Yuan-Fang and Pan, Shirui and Wen, Qingsong},
  booktitle = {The Twelfth International Conference on Learning Representations},
  year      = {2024},
  url       = {https://openreview.net/forum?id=Unb5CVPtae}
}

@article{zhou2023onefitsall,
  title={One fits all: Power general time series analysis by pretrained lm},
  author={Zhou, Tian and Niu, Peisong and Sun, Liang and Jin, Rong and others},
  journal={Advances in neural information processing systems},
  volume={36},
  pages={43322--43355},
  year={2023}
}

@inproceedings{wang2025chattime,
  title     = {{ChatTime}: A Unified Multimodal Time Series Foundation Model Bridging Numerical and Textual Data},
  author    = {Wang, Chengsen and Qi, Qi and Wang, Jingyu and Sun, Haifeng and Zhuang, Zirui and Wu, Jinming and Zhang, Lei and Liao, Jianxin},
  booktitle = {Proceedings of the AAAI Conference on Artificial Intelligence},
  volume    = {39},
  number    = {12},
  pages     = {12694--12702},
  year      = {2025},
  doi       = {10.1609/aaai.v39i12.33384}
}

@article{romeraparedes2024funsearch,
  title   = {Mathematical Discoveries from Program Search with Large Language Models},
  author  = {Romera-Paredes, Bernardino and Barekatain, Mohammadamin and Novikov, Alexander and Balog, Matej and Kumar, M. Pawan and Dupont, Emilien and Ruiz, Francisco J. R. and Ellenberg, Jordan S. and Wang, Pengming and Fawzi, Omar and Kohli, Pushmeet and Fawzi, Alhussein},
  journal = {Nature},
  volume  = {625},
  pages   = {468--475},
  year    = {2024},
  doi     = {10.1038/s41586-023-06924-6}
}

@article{bran2024chemcrow,
  title   = {Augmenting Large Language Models with Chemistry Tools},
  author  = {Bran, Andres M. and Cox, Sam and Schilter, Oliver and Baldassari, Carlo and White, Andrew D. and Schwaller, Philippe},
  journal = {Nature Machine Intelligence},
  volume  = {6},
  pages   = {525--535},
  year    = {2024},
  doi     = {10.1038/s42256-024-00832-8}
}

@inproceedings{sprueill2024chemreasoner,
  title     = {{CHEMREASONER}: Heuristic Search over a Large Language Model's Knowledge Space Using Quantum-Chemical Feedback},
  author    = {Sprueill, Henry W. and Edwards, Carl and Agarwal, Khushbu and Olarte, Mariefel V. and Sanyal, Udishnu and Johnston, Conrad and Liu, Hongbin and Ji, Heng and Choudhury, Sutanay},
  booktitle = {Proceedings of the 41st International Conference on Machine Learning},
  series    = {Proceedings of Machine Learning Research},
  volume    = {235},
  pages     = {46351--46374},
  year      = {2024},
  publisher = {PMLR}
}

@inproceedings{yang2024opro,
  title     = {Large Language Models as Optimizers},
  author    = {Yang, Chengrun and Wang, Xuezhi and Lu, Yifeng and Liu, Hanxiao and Le, Quoc V. and Zhou, Denny and Chen, Xinyun},
  booktitle = {The Twelfth International Conference on Learning Representations},
  year      = {2024}
}

@inproceedings{ma2024eureka,
  title     = {{Eureka}: Human-Level Reward Design via Coding Large Language Models},
  author    = {Ma, Yecheng Jason and Liang, William and Wang, Guanzhi and Huang, De-An and Bastani, Osbert and Jayaraman, Dinesh and Zhu, Yuke and Fan, Linxi and Anandkumar, Anima},
  booktitle = {The Twelfth International Conference on Learning Representations},
  year      = {2024},
  url       = {https://openreview.net/forum?id=IEduRUO55F}
}

@article{gottweis2026coscientist,
  title={Accelerating scientific discovery with Co-Scientist},
  author={Gottweis, Juraj and Weng, Wei-Hung and Daryin, Alexander and Tu, Tao and Sirkovic, Petar and Myaskovsky, Artiom and Glowaty, Grzegorz and Weissenberger, Felix and Orlandi, Alessio and Popovici, Dan and others},
  journal={Nature},
  volume={655},
  number={8122},
  pages={487--496},
  doi={10.1038/s41586-026-10644-y},
  year={2026},
  publisher={Nature Publishing Group UK London}
}

\clearpage
\appendix

\makeatletter
\setlength{\@fptop}{0pt}
\setlength{\@fpsep}{12pt}
\setlength{\@fpbot}{0pt plus 1fil}
\makeatother
\renewcommand{\topfraction}{0.95}
\renewcommand{\bottomfraction}{0.95}
\renewcommand{\textfraction}{0.05}
\renewcommand{\floatpagefraction}{0.80}
\setlength{\textfloatsep}{10pt plus 2pt minus 2pt}
\setlength{\floatsep}{10pt plus 2pt minus 2pt}
\setlength{\emergencystretch}{2em}
\newcolumntype{Z}{>{\fontsize{0.1pt}{0.1pt}\selectfont}r}

\section{Experimental Protocol and Evidence Interfaces}
\label{sec:appendix}

\sysname denotes the complete workflow with candidate search, Criteria-to-Measurement Compilation, Evidence-Grounded Failure Review, independent subagent execution, regression checks, governed candidate promotion, and a frozen final predictor.

\subsection{Evaluation Protocol and Data Splits}
\label{app:protocol}

All experiments evaluate multi-label diagnosis from 12-lead ECGs. Methods use the same label mapping, lead order, preprocessing, validation-based model selection, and validation-based threshold selection within each dataset. Test sets are used only for final reporting. Unless otherwise stated, \sysname uses DeepSeek-V4-Pro, candidate count $k=3$, maximum iteration count 5.
This subsection expands the compact experimental setup in \secref{sec:datasets}: \tabref{tab:app_dataset_scope} gives the evaluation scale, and \tabref{tab:app_dataset_preprocessing} specifies the full label sets and preprocessing choices that define the fixed problem contracts.

\begin{table*}[t]
\centering
\caption{Dataset and evaluation scope.}
\label{tab:app_dataset_scope}
\small
\setlength{\tabcolsep}{6pt}
\begin{tabularx}{0.92\textwidth}{lrrr>{\raggedright\arraybackslash}X}
\toprule
Dataset & Train & Validation & Test & Evaluated labels \\
\midrule
PTB-XL & 17,441 & 2,193 & 2,203 & 5 diagnostic superclasses; 23 diagnostic subclasses for fine-grained analysis \\
Georgia & 4,946 & 581 & 599 & NSR, AF, IAVB, LBBB, RBBB, SB, STach \\
CPSC2018 & 4,126 & 1,375 & 1,376 & Normal/NSR, AF, IAVB, LBBB, RBBB, PAC, PVC, STD, STE \\
\bottomrule
\end{tabularx}
\end{table*}

The three datasets cover different label organizations. PTB-XL provides a superclass/subclass diagnostic hierarchy, Georgia uses a seven-label scored setting, and CPSC2018 is treated as a nine-label arrhythmia task under a fixed 60/20/20 split. This makes Macro F1 the primary metric because the label supports are highly imbalanced and several clinically meaningful labels are rare.
All reported standard deviation is computed with a denominator of $N$ (i.e., \texttt{ddof=0}).

\begin{table*}[t]
\centering
\caption{Dataset-specific labels and preprocessing used in the experiments.}
\label{tab:app_dataset_preprocessing}
\small
\renewcommand{\arraystretch}{1.05}
\setlength{\tabcolsep}{3.5pt}
\begin{tabularx}{0.98\textwidth}{>{\raggedright\arraybackslash}p{0.13\textwidth}>{\raggedright\arraybackslash}p{0.28\textwidth}>{\raggedright\arraybackslash}p{0.21\textwidth}>{\raggedright\arraybackslash}X}
\toprule
Dataset & Complete evaluated label set & Split protocol & Preprocessing \\
\midrule
PTB-XL &
Main task: NORM, MI, STTC, CD, HYP. Fine-grained analysis: NORM, IMI, AMI, LMI, PMI, STTC, NST\_, ISC\_, ISCA, ISCI, LAFB/LPFB, IRBBB, \_AVB, IVCD, CRBBB, CLBBB, WPW, ILBBB, LVH, LAO/LAE, RVH, RAO/RAE, SEHYP. &
Official folds: 1--8 for training, 9 for validation, and 10 for testing. &
Official 100 Hz 12-lead waveforms are used in canonical lead order. Signals are loaded as float32, non-finite values are set to zero, and each lead is normalized with training-set statistics before model fitting and evaluation. Records keep the official 10 s duration. \\
\midrule
Georgia &
NSR, AF, IAVB, LBBB, RBBB, SB, STach. &
Record groups 3--11 for training, group 2 for validation, and group 1 for testing. &
Header metadata and SNOMED codes are mapped to the seven scored labels used by the baseline protocol. Signals are converted to physical units, reordered to canonical 12-lead order, resampled to 100 Hz, padded or cropped to 10 s, cleaned for missing or non-finite values, and normalized using training-set statistics. \\
\midrule
CPSC2018 &
Normal/NSR, AF, IAVB, LBBB, RBBB, PAC, PVC, STD, STE. &
Fixed record-level 60/20/20 train/validation/test split with primary-label stratification. &
SNOMED codes are mapped to the nine target labels. Signals are converted to physical units, reordered to canonical 12-lead order, resampled to 250 Hz, padded or cropped to 10 s, cleaned for missing or non-finite values, and normalized with statistics computed from the training split. \\
\bottomrule
\end{tabularx}
\renewcommand{\arraystretch}{1.0}
\end{table*}

\begin{table*}[t]
\centering
\caption{Default \sysname configuration used for the main experiments.}
\label{tab:app_sysname_config}
\small
\renewcommand{\arraystretch}{1.08}
\setlength{\tabcolsep}{4pt}
\begin{tabularx}{0.96\textwidth}{>{\raggedright\arraybackslash}p{0.18\textwidth}>{\raggedright\arraybackslash}X}
\toprule
Component & Setting \\
\midrule
Offline controller & DeepSeek-V4-Pro for development-time planning, evidence review, and candidate proposal; no LLM calls at test time. \\
Candidate budget & $k=3$ executable candidate revisions per iteration. \\
Iteration budget & Maximum of five refinement iterations, unless explicitly changed in a sensitivity study. \\
Data access & Train for fitting, validation for model/threshold selection and promotion, and test only for final reporting. \\
Promotion objective & Promote only when validation score improves over the incumbent and execution, reproducibility, and regression checks pass. \\
Thresholds and runs & Label-wise validation thresholds; metrics reported as mean~$\pm$~standard deviation over five runs. \\
Audit artifacts & Problem contract, measurement interface, selected and comparator cases, candidate logs, promotion decisions, and final outputs. \\
\bottomrule
\end{tabularx}
\renewcommand{\arraystretch}{1.0}
\end{table*}

\tabref{tab:app_sysname_config} summarizes the default configuration used to instantiate the workflow in \figref{fig:method_overview}. These settings are held fixed for the main comparison unless a later appendix table explicitly studies a sensitivity setting, such as the candidate budget or LLM backbone.
\tabref{tab:app_measurement_inventory} lists the measurement interface produced by one representative CMC run. The table is not a new diagnostic rule set; it is an inventory of the deterministic evidence channels available to EGFR. The WPW example in \figref{fig:case_wpw_pseudo_mi} and \tabref{tab:app_wpw_case_review} shows how several of these functions are used in an individual failure review.

\begin{table*}[t]
\centering
\caption{Measurement-function inventory from a representative \sysname run. Each function emits a numeric value or structured status together with its source evidence pointer and applicability flag.}
\label{tab:app_measurement_inventory}
\small
\renewcommand{\arraystretch}{1.02}
\setlength{\tabcolsep}{4pt}
\begin{tabularx}{0.96\textwidth}{>{\raggedright\arraybackslash}p{0.26\textwidth}>{\raggedright\arraybackslash}X}
\toprule
Measurement function & Emitted evidence and review use \\
\midrule
Signal quality and lead availability & Missing leads, non-finite values, flat segments, and usable waveform length; used to mark unreliable evidence. \\
R-peak and beat localization & Detected beat times, representative beat, and confidence; used by rhythm and interval measurements. \\
Heart-rate summary & Mean heart rate and RR statistics; used for bradycardia, tachycardia, and rhythm plausibility. \\
Rhythm regularity & RR variability and irregularity status; used for AF, premature-beat patterns, and comparator selection. \\
PR interval & PR duration and short/long PR status; used for IAVB and pre-excitation evidence. \\
QRS duration & QRS width and wide-QRS status; used for bundle-branch block and ventricular-conduction evidence. \\
QT and corrected QT & QT/QTc estimates with unsupported flags when landmarks are unreliable; used for repolarization review. \\
Frontal QRS axis & Limb-lead polarity and axis-quadrant summary; used for conduction and hypertrophy evidence. \\
ST-segment deviation & Lead-wise ST elevation/depression summaries; used for STTC, STD, STE, ischemia, and infarction-like failures. \\
T-wave morphology & T-wave polarity and repolarization cues; used for ST--T subclasses and comparator evidence. \\
Pathological Q-wave screen & Initial deflection and Q-wave-like morphology by lead group; used for MI, pseudo-infarction, and WPW review. \\
R-wave progression & Precordial R/S progression and transition pattern; used for MI, conduction, and hypertrophy-related failures. \\
Bundle-branch morphology & V1--V2 and lateral-lead terminal-force patterns; used for LBBB/RBBB failure review. \\
AV-conduction screen & PR prolongation and conduction-delay status; used for IAVB and AV-block-related errors. \\
Pre-excitation screen & Short PR, slurred QRS onset, widened QRS, and axis consistency; used for WPW and pseudo-infarction cases. \\
Voltage hypertrophy screen & Limb/precordial voltage summaries and cross-lead high-voltage pattern; used for LVH/RVH and HYP errors. \\
Premature-beat evidence & Early-beat and compensatory-pause cues when detectable; used for PAC, PVC, and rhythm-label failures. \\
Cross-lead consistency summary & Agreement or conflict among local morphology, rhythm, and axis evidence; used by EGFR for comparator construction and revision hypotheses. \\
\bottomrule
\end{tabularx}
\renewcommand{\arraystretch}{1.0}
\end{table*}

\tabref{tab:app_measurement_inventory} is intentionally reported at the function-family level rather than as source code. The validation procedure, artifact requirements, and deterministic execution contract for these functions are detailed next in \secref{app:reference_corpus} and \secref{app:extractor_validation}.

\subsection{Prompt Templates and Control Instructions}
\label{app:prompt_templates}

\tabref{tab:app_prompt_templates} reports the key prompt excerpts
used by the agent calls in \secref{sec:criteria_measurement},
\secref{sec:failure_review}, and \secref{sec:classifier_revision}.
We report only the operationally important parts of each prompt,
because the full runtime prompt also includes workspace paths, current
iteration state, available tools, and task-specific attachments.

\begin{table*}[t]
\centering
\caption{Key prompt excerpts used by the main RecursiveECG agent calls.}
\label{tab:app_prompt_templates}
\small
\setlength{\tabcolsep}{3pt}
\begin{tabularx}{\textwidth}{>{\raggedright\arraybackslash}p{0.13\textwidth}>{\raggedright\arraybackslash}p{0.24\textwidth}>{\raggedright\arraybackslash}X}
\toprule
Symbol / prompt & Role & Key prompt excerpt \\
\midrule
\texttt{p\_cmp} &
Criteria-to-Measurement Compilation &
Generate a deterministic and auditable reference feature extractor from the current task definition and fixed ECG references. Read the problem contract, data specification, and listed references first. Do not add thresholds, diagnostic rules, or feature definitions from model common sense. Every feature, computation method, threshold, and judgment must be traceable to reference text; unsupported or unobservable conditions must return \texttt{indeterminate}. Write an evidence map and measurement plan before implementing the extractor, then validate the Python API and CLI on at least one real workspace ECG sample. \\
\midrule
\texttt{p\_diag} &
Evidence-Grounded Failure Review &
Use validation feedback, selected failed cases, correctly classified comparator cases, model outputs, raw ECG waveforms, and reference-backed measurements to diagnose why the current predictor fails. Compare failed cases with clinically similar successful cases. Base each important judgment on data observations, tool outputs, or reference-backed evidence. If the bad case cannot be explained by numerical or clinical evidence, explicitly state that the evidence is insufficient rather than forcing a domain explanation. Output a concise, evidence-supported classifier limitation for the next revision. \\
\midrule
\texttt{p\_rev} &
Diagnostic Classifier Revision &
Given the current predictor implementation and the diagnosed limitation, generate executable code-level revisions rather than natural-language suggestions. Preserve the fixed label space, data split, decision rule, evaluation metric, leakage constraints, and validation protocol. Execute and evaluate each candidate under the same training and validation protocol. Reject candidates that fail to run, violate the contract, leak test information, or degrade validation performance. Retain a revision only when executable validation evidence supports promotion. \\
\midrule
Main orchestrator &
Node routing and workflow control &
Use the backend-provided workspace progress as the source of truth for routing. Enter the recommended node only when the workflow should continue; do not perform substantive node work in the main session. Query the knowledge graph, reference QA, or extractor-source QA only when the corresponding tool is explicitly available. Do not start a second node while another node is active. \\
\midrule
Node execution and finish control &
Reproducible node execution &
Complete only the current node's responsibility and cite background knowledge, data observations, or tool outputs for important judgments. Use only node-native tools and injected Harness tools. Avoid broad process-killing commands. At completion, call the finish-control tool with success status, a short summary, key artifact paths, and, for iterative solving, an explicit continue-or-exit decision. \\
\bottomrule
\end{tabularx}
\end{table*}

\subsection{Reference Corpus and Evidence Binding}
\label{app:reference_corpus}

The reference corpus $\mathcal{K}$ is a fixed development-time artifact rather than a source that is searched during deployment-time inference.
For each dataset, it contains the task contract, dataset label definitions, and ECG interpretation references relevant to the target label space.
For the ECG criteria used to construct the reference-backed measurement functions and support Evidence-Grounded Failure Review, we use standard ECG interpretation material covering ECG acquisition and intervals, diagnostic statement terminology, intraventricular conduction, ST-segment and repolarization patterns, chamber- and axis-related changes, and acute ischemia/infarction morphology~\cite{kligfield2007ecgpart1,mason2007ecgpart2,surawicz2009ecgpart3,rautaharju2009ecgpart4,hancock2009ecgpart5,wagner2009ecgpart6}.
These references support the construction of auditable measurement functions; they are not used to claim clinical guideline validation of the final predictor.

Reference selection follows three constraints.
First, the corpus is frozen before iterative solving starts, and the test split is never used to select or modify reference criteria.
Second, a candidate measurement must be linked to the current problem contract, label set, or a concrete need identified during Evidence-Grounded Failure Review; generic ECG knowledge that cannot affect the task is not compiled into the measurement interface.
Third, every compiled measurement function must carry an evidence record containing a reference path, section or page pointer, and a concise description of the operationalized rule.
If the reference material or waveform quality does not support a reliable judgment for an input condition, the measurement function remains available but emits an unsupported or indeterminate status rather than a speculative normal/abnormal judgment.

This evidence-binding mechanism is designed to prevent knowledge drift across iterations.
During candidate generation and Evidence-Grounded Failure Review, the agent may inspect the emitted measurement records and their linked evidence, but it does not repeatedly reinterpret unconstrained clinical text.
The same reference-backed computation is therefore reused across candidate testing, failure attribution, case replay, and final audit reporting.

\subsection{Failure Selection and Comparator Retrieval}
\label{app:failure_retrieval}

The failure-review stage is executed only on the validation split and is
never allowed to inspect test labels. We make the selection and retrieval
contract explicit here because aggregate validation loss alone is not a
sufficient description of the cases used to formulate a revision.
For a record $i$ and label $k$, let $y_{ik}$ be the ground-truth label,
$p_{ik}$ the predicted probability, and $\widehat{y}_{ik}$ the thresholded
decision. The per-label binary cross-entropy contribution is
\[
\ell_{ik}=-y_{ik}\log(p_{ik}+\epsilon)
-(1-y_{ik})\log(1-p_{ik}+\epsilon),
\]
and the per-record error score is
\[
u_i=\frac{1}{L}\sum_{k=1}^{L}\ell_{ik}.
\]
We quantify the influence of label $k$ on the current failure pool by
\[
I_k=\frac{\sum_{i\in\mathcal{D}_{\mathrm{val}}}
\mathbf{1}[\widehat{y}_{ik}\ne y_{ik}]\ell_{ik}}
{\sum_{j=1}^{L}\sum_{i\in\mathcal{D}_{\mathrm{val}}}
\mathbf{1}[\widehat{y}_{ij}\ne y_{ij}]\ell_{ij}+\epsilon}.
\]
Thus, failure influence is not a subjective label importance score: it is
the fraction of error-weighted validation loss attributable to a label.
The reported error-mode taxonomy is the Cartesian product of the affected
label set and the decision error type: false negative, false positive, or
mixed error. We additionally mark a case as persistent when it recurs for
the incumbent across consecutive iterations, and as borderline when the
decision margin is within the pre-specified review band around the label
threshold.

The selection procedure has two stages. First, it forms a failure pool
$\mathcal{D}^{-}_{\mathrm{val},\theta}$ and assigns each case its $u_i$,
affected-label set, error type, persistence flag, and measurement-quality
status. Second, it samples within error-mode strata, prioritizing the
largest $u_i$ cases and retaining at least one case for each targeted
failure mode whenever that mode is present. If a stratum contains fewer
cases than its quota, the unused quota is assigned to the next highest
influence stratum. Correctly classified cases are sampled as controls and
are never mixed into the failure pool.

The concrete audit traces use a small, fixed review budget. PTB-XL reviews
use five bad cases and two correctly classified controls per iteration in
the final case-review trace; CPSC2018 uses stratified samples of 12--15
bad cases when the validation failure pool is large; and the Georgia audit
uses 15 stratified cases from the 86-case error pool in its available
case-review trace. These counts describe review artifacts, not the number
of cases used to train or evaluate the classifier.

For comparator retrieval, the candidate pool is
\[
\mathcal{D}^{+}_{\mathrm{val},\theta}=
\{(X_j,\mathbf{y}_j):\widehat{\mathbf{y}}_j=\mathbf{y}_j\},
\]
restricted to the same dataset and split and excluding the failure record
itself. A comparator is ranked by the following deterministic evidence
key, in descending lexicographic order:
\begin{enumerate}
\item label-profile agreement, measured by Jaccard overlap of the
      ground-truth label sets and agreement on the relevant error-mode
      labels;
\item measurement-profile agreement, using exact agreement for categorical
      statuses and robustly normalized distance for numeric measurements
      that are available for both cases;
\item waveform-profile agreement, using lead-wise normalized morphology
      summaries derived from the same 12-lead input, including rhythm,
      QRS, ST--T, axis, and voltage summaries; and
\item deterministic record-ID order as a tie-breaker.
\end{enumerate}
Only correctly classified cases are eligible. We retain at most two
comparators per failure stratum to prevent a large class from dominating
the diagnosis. A missing measurement is not treated as a match: it is
excluded from the numeric distance and retained as an explicit
\texttt{indeterminate} status. If no eligible comparator satisfies the
label and evidence filters, the case is retained with the status
\texttt{no-comparator}; EGFR may use its raw waveform and measurement
records, but it must not infer a contrastive mechanism from an absent
reference case. This fallback is reported as an evidence limitation rather
than silently replacing retrieval with unconstrained LLM judgment.

The resulting provenance is summarized in
\tabref{tab:app_failure_selection_audit}. It makes the split, incumbent,
failure-pool size, reviewed-case budget, and stratification rule explicit
for each available case-review trace, so that the subsequent
evidence-grounded analysis can be interpreted as a development-time audit
rather than as an additional evaluation protocol.

\begin{table*}[t]
\centering
\caption{Failure-selection audit traces from the available case-review workspaces. The table reports the split and incumbent used by each trace; it is not a cross-dataset performance comparison.}
\label{tab:app_failure_selection_audit}
\small
\setlength{\tabcolsep}{4pt}
\begin{tabularx}{0.96\textwidth}{lrrrr>{\raggedright\arraybackslash}X}
\toprule
Dataset / trace & Evaluation records & Failure pool & Bad cases reviewed & Controls & Sampling rule \\
\midrule
PTB-XL / final review trace & 2,193 validation & 862 & 5 & 2 & Stratified by HYP false negatives, persistent failures, borderline cases, and multi-label confusion. \\
Georgia / validation case-review trace & 581 validation & 86 & 15 & 6 & Stratified across AF false negatives, NSR/IAVB false positives, multi-label errors, and diverse edge cases; extractor unavailable in this trace. \\
CPSC2018 / final review trace & 1,375 validation & 300 & 15 & 5 & Stratified by error type and severity, with coverage of PAC, RBBB, PVC, STD, and STE failure modes. \\
\bottomrule
\end{tabularx}
\end{table*}

The evaluation-record counts follow the official dataset split sizes
reported in the experimental setup. The failure-pool and reviewed-case
counts are the corresponding case-review trace statistics after the
trace-specific filtering and sampling procedure. The Georgia row is a
validation-set case-review trace: it analyzes the 581-record validation
split used by the development protocol. It therefore belongs to the
development-time refinement evidence and is distinct from the held-out
test split, which remains reserved for final reporting. The
row is included to document the provenance and limits of the validation
case-review artifact.

\subsection{Measurement Function Construction and Validation}
\label{app:extractor_validation}

The Criteria-to-Measurement Compilation stage builds the deterministic measurement interface as a set of validated executable artifacts under a fixed input-output contract.
The implementation requires an evidence map, measurement plan, Python measurement extractor, manifest, reference-rule file, README, test cases, and evaluation report before the interface can be used downstream.
The backend validator then checks that these artifacts are complete, mutually consistent, deterministic, executable, and grounded in the cited reference evidence.

The validator checks are listed in \tabref{tab:extractor_validation}.
They define the minimum artifact, evidence-binding, determinism, execution,
and control-case requirements that must be satisfied before emitted
measurements are used by Evidence-Grounded Failure Review. The resulting
dataset-level status counts are reported separately in
\tabref{tab:app_extractor_validation_results}.

\begin{table*}[t]
\centering
\caption{Validation checks for deterministic measurement functions.}
\label{tab:extractor_validation}
\small
\setlength{\tabcolsep}{4pt}
\begin{tabularx}{0.94\textwidth}{>{\raggedright\arraybackslash}p{0.24\textwidth}>{\raggedright\arraybackslash}X}
\toprule
Check & Requirement \\
\midrule
Schema completeness & The manifest must declare the entrypoint, Python API, input and output schemas, and non-empty measurement definitions. \\
Evidence binding & Each manifest measurement, evidence-map entry, and reference rule must contain reference evidence with a path and section or page pointer. \\
Rule consistency & Measurement names must match across the manifest, measurement plan, evidence map, and reference-rule file. \\
Deterministic source & The measurement extractor cannot use network calls, random number generators, system time, external processes, or hidden mutable state. \\
Execution consistency & The importable Python API and command-line wrapper must return the same JSON object on the same input. \\
Repeatability & Repeated execution on the same JSON input must produce identical output. \\
Real-sample behavior & At least one test case must be extracted from a real workspace sample following the data specification. \\
Control behavior & The evaluation report must include control or reference cases and record abnormal burden, warnings, unsupported conditions, and indeterminate measurements. \\
\bottomrule
\end{tabularx}
\end{table*}

\begin{table*}[t]
\centering
\caption{Quantitative validation report for the reference-backed measurement extractors. Counts are aggregated over emitted feature records; \texttt{descriptive} denotes a non-judgmental signal summary and is therefore excluded from the normal/abnormal/indeterminate denominator.}
\label{tab:app_extractor_validation_results}
\small
\setlength{\tabcolsep}{4pt}
\begin{tabularx}{0.96\textwidth}{lrrrrr>{\raggedright\arraybackslash}X}
\toprule
Dataset & Cases & Control cases & Normal & Abnormal & Indeterminate & Observed limitation \\
\midrule
PTB-XL & 2 & 1 & 16 & 16 & 0 & 3/16 abnormal judgments on the NORM control; QTc, T-wave discordance, and pathological-Q detection are sensitive to 100 Hz resolution. \\
Georgia & 7 & 1 & 27 & 13 & 9 & The NSR control has 1 abnormal and 3 indeterminate judgments out of 7; P-wave and area-based axis heuristics are unreliable. \\
CPSC2018 & 28 & 2 & 144 & 196 & 24 & The report additionally contains 84 descriptive summaries; all judgments are deterministic heuristics rather than clinical diagnoses. \\
\bottomrule
\end{tabularx}
\end{table*}

The validation reports provide three distinct kinds of evidence. First,
the extractor is executable and deterministic: each dataset report is
generated from fixed JSON cases and the extractor contract forbids network
access, randomness, hidden mutable state, and model-dependent computation.
Second, the status counts expose uncertainty instead of converting failed
landmark detection into a normal or abnormal decision. Third, the control
cases reveal dataset- and sampling-specific failure modes. In particular,
the PTB-XL report attributes three control abnormalities to 100 Hz
time-resolution limitations, the Georgia report identifies P-wave and
axis weaknesses, and the CPSC2018 report records indeterminate QRS/PR
measurements together with non-judgmental descriptive features. These
results support the claim that the functions are executable, traceable,
and useful for model-development evidence; they do not establish
expert-level clinical measurement accuracy.

The construction workflow is intentionally conservative.
The agent first writes an evidence map and measurement plan before writing executable code.
The evidence map binds each proposed measurement to its source reference, while the measurement plan records its unit, computation procedure, judgment rules, control expectation, uncertainty conditions, and expected failure modes.
Only after these planning artifacts exist does the agent implement the measurement extractor.
The extractor receives all waveform data through explicit JSON input and is not allowed to read raw data files, reference files, network resources, or model state during execution.

Validation uses both artifact checks and real-sample execution checks.
For each test case, the backend executes the measurement extractor twice and compares the complete JSON outputs.
When command-line execution is enabled, it also compares the Python module API against the command-line interface.
A real sample is required so that validation is not limited to synthetic smoke tests.
The evaluation report summarizes measurement-status counts for real and control cases, including warnings, unsupported states, and indeterminate outputs.
If a control case produces an unexpectedly high abnormal burden, the corresponding measurement function must be debugged, weakened to an advisory cue, or marked with an explicit limitation before Evidence-Grounded Failure Review can use its output as case-level evidence.
Thus, the measurement interface is validated as a reproducible evidence interface rather than as an independent clinical diagnostic system.

\subsection{Reference-Backed Measurement Interface Example}
\label{app:feature_extractor_example}

Algorithm~\ref{alg:feature_extractor_example} gives a compact description of the deterministic reference-backed measurement interface used during Evidence-Grounded Failure Review.
The example omits low-level signal-processing details but shows the auditable output contract: each measurement function emits a measured value, unit, judgment, operational rule, and reference evidence pointer.
The interface is deterministic and receives all waveform data as explicit JSON input; it does not call an LLM, train a model, or read external state during either development-time replay or deployment-time inference.

\begin{algorithm}[t]
\caption{Reference-backed deterministic measurement emission}
\label{alg:feature_extractor_example}
\KwIn{ECG case $X$ with 12-lead waveform and sampling rate; fixed measurement-rule registry $\mathcal{R}$}
\KwOut{Measurement report $E(X)$ with measured values, judgments, and evidence pointers}
Initialize $E(X) \leftarrow \emptyset$\;
\ForEach{registered measurement criterion $r \in \mathcal{R}$}{
    Select the required leads, fiducial points, and measurement window specified by $r$\;
    Compute the numeric measurement $v_r$ using deterministic signal-processing routines\;
    \eIf{$v_r$ is unavailable or the required waveform support is invalid}{
        Set judgment $q_r \leftarrow$ indeterminate\;
    }{
        Compare $v_r$ with the threshold or predicate encoded in $r$ and set $q_r \leftarrow$ normal or abnormal\;
    }
    Append the measurement record
    $\{name:r.name,\ value:v_r,\ unit:r.unit,\ judgment:q_r,\ rule:r.text,\ evidence:r.references\}$
    to $E(X)$\;
}
\Return{$E(X)$}\;
\end{algorithm}

\section{Additional Diagnostic Results}
\label{app:diagnostic_results}

\subsection{Aggregate Diagnostic Performance}
\label{app:aggregate}

The primary aggregate result is that \sysname{} improves macro F1
over the strongest baseline on all three datasets. The absolute gain
is 0.0403 on PTB-XL, 0.0696 on Georgia, and 0.1096 on CPSC2018.

The aggregate results support a focused but important claim.
\sysname{} does not improve only one benchmark or one label
organization. It improves the primary metric under the PTB-XL
diagnostic hierarchy, the Georgia scored-label setting, and the
CPSC2018 arrhythmia label set. The largest absolute gain occurs on
CPSC2018. However, this dataset also contains the strongest
label-level F1 exception, RBBB, which is examined below. Thus, the
aggregate improvement should be interpreted together with the
fine-grained positive and negative results rather than as uniform
per-label dominance.

\subsection{Dataset-Adaptive Architecture Evolution}
\label{app:architecture_transfer_latest}

The architecture-transfer experiment asks whether the architecture
and governed toolchain structure discovered on PTB-XL can be reused
on new ECG datasets. This setting transfers architecture and
toolchain design rather than trained weights or zero-shot
predictions. The target output layer is replaced and retrained, but
the agent workflow is not rerun on the target dataset.

\begin{table*}[t]
\centering
\caption{Architecture transfer from PTB-XL to target datasets under
the transfer-control protocol.}
\label{tab:app_arch_transfer_latest}
\small
\setlength{\tabcolsep}{5pt}
\begin{tabularx}{0.90\textwidth}{>{\raggedright\arraybackslash}Xlccc}
\toprule
Architecture & Target & Target workflow rerun & Macro F1 & Gap \\
\midrule
PTB-XL-derived & Georgia & No & 0.8368 & 0.0458 \\
Target-specific \sysname{} & Georgia & Yes & 0.8826 & 0.0000 \\
PTB-XL-derived & CPSC2018 & No & 0.7011 & 0.1016 \\
Target-specific \sysname{} & CPSC2018 & Yes & 0.8027 & 0.0000 \\
\bottomrule
\end{tabularx}
\end{table*}

The transferred architecture remains competitive, but it does not
match the corresponding target-specific rerun under the transfer
protocol. This result indicates that \sysname{} discovers reusable
ECG model structures, but that architecture reuse alone does not
recover the complete benefit observed under target-specific refinement.

In particular, target-dataset Evidence-Grounded Failure Review
(EGFR), the execution and use of reference-backed measurements
produced through Criteria-to-Measurement Compilation (CMC), and
governed toolchain revision expose label-set-specific and
distribution-specific failure patterns that the PTB-XL-derived
architecture alone cannot repair. The transfer result therefore
separates reusable architecture discovery from target-specific
evidence-grounded refinement under this protocol.

\subsection{Fine-Grained Label Behavior}
\label{app:fine_grained_latest}

Fine-grained analysis shows that the aggregate gains are not
uniform, and that F1 and AUC provide complementary views of model
behavior. In terms of F1, \sysname{} improves many
morphology-sensitive and rhythm-sensitive labels, particularly the
PTB-XL ST--T and conduction subclasses and the CPSC2018 PAC, STE,
STD, NSR, and AF labels. The principal F1 exceptions are PTB-XL RVH
and SEHYP and CPSC2018 RBBB.

The complete per-label F1 results are reported in
\tabref{tab:app_ptb_subclass_latest}, \tabref{tab:app_georgia_label_latest},
and \tabref{tab:app_cpsc_label_latest}; the corresponding AUC results are
reported in \tabref{tab:ptbxl-subclass-auc}, \tabref{tab:georgia-label-auc},
and \tabref{tab:cpsc2018-label-auc}. These tables provide the numerical
breakdown behind the aggregate and exception patterns described above,
with deltas computed against the strongest baseline for each label.
It is worth noting that, since PTB-XL provides an explicit 23-subclass classification setting, the PTB-XL results here are obtained by replacing the classification head with a 23-class classification head and recomputing the metrics.

The AUC results show a broader improvement in ranking quality.
\sysname{} improves over the strongest baseline on 17 of the 23
PTB-XL subclasses, ties on ISC\_ and CLBBB, improves on 6 of the 7
Georgia labels, and improves on 8 of the 9 CPSC2018 labels. The remaining
AUC regressions are PTB-XL CRBBB, ILBBB, LAFB/LPFB, and SEHYP,
together with Georgia RBBB and CPSC2018 RBBB.

\begin{figure*}[t]
\centering
\includegraphics[width=0.9\textwidth]{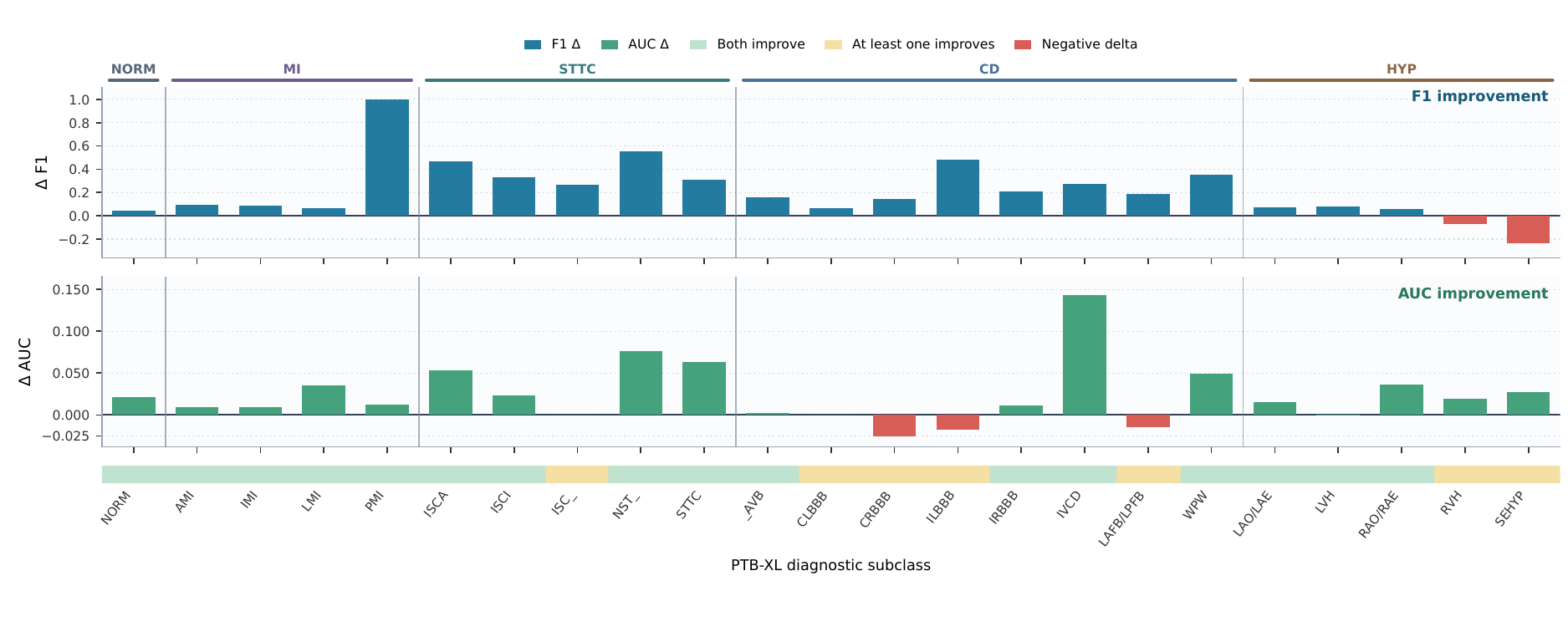}
\caption{Fine-grained F1 and AUC changes of \sysname{} relative to
the strongest baseline on PTB-XL.}
\Description{Two grouped plots showing per-subclass F1 and AUC
differences between RecursiveECG and the strongest baseline on
PTB-XL.}
\label{fig:app_label_delta_latest}
\end{figure*}

\begin{table*}[t]
\centering
\caption{PTB-XL diagnostic subclass F1 results reported as mean~$\pm$~standard deviation over five runs. Delta is computed from mean values.}
\label{tab:app_ptb_subclass_latest}
\scriptsize
\setlength{\tabcolsep}{3pt}
\resizebox{\textwidth}{!}{%
\begin{tabular}{llrrrrrrrrrlr}
\toprule
Superclass & Subclass & Support & PatchTST & TimesNet & TS2Vec & xLSTM & UniTS & MERL & Argos & \sysname & Best baseline & Delta \\
\midrule
NORM & NORM & 957 & $0.7104{\pm}0.0049$ & $0.7106{\pm}0.0076$ & $0.7592{\pm}0.0052$ & $0.8301{\pm}0.0057$ & $0.8037{\pm}0.0068$ & $0.7931{\pm}0.0104$ & $0.6818{\pm}0.0002$ & $0.8757{\pm}0.0048$ & xLSTM & 0.0456 \\
MI & AMI & 307 & $0.5907{\pm}0.0078$ & $0.5385{\pm}0.0120$ & $0.7568{\pm}0.0063$ & $0.6182{\pm}0.0047$ & $0.6783{\pm}0.0065$ & $0.6967{\pm}0.0052$ & $0.4286{\pm}0.0026$ & $0.8502{\pm}0.0051$ & TS2Vec & 0.0934 \\
MI & IMI & 330 & $0.5025{\pm}0.0036$ & $0.4214{\pm}0.0086$ & $0.6762{\pm}0.0087$ & $0.6529{\pm}0.0116$ & $0.6078{\pm}0.0059$ & $0.6478{\pm}0.0095$ & $0.4035{\pm}0.0027$ & $0.7606{\pm}0.0079$ & TS2Vec & 0.0844 \\
MI & LMI & 20 & $0.0625{\pm}0.0182$ & $0.0000{\pm}0.0000$ & $0.1395{\pm}0.0140$ & $0.0800{\pm}0.0107$ & $0.1875{\pm}0.0097$ & $0.1026{\pm}0.0089$ & $0.0000{\pm}0.0000$ & $0.2500{\pm}0.0208$ & UniTS & 0.0625 \\
MI & PMI & 2 & $0.0000{\pm}0.0000$ & $0.0000{\pm}0.0000$ & $0.0000{\pm}0.0000$ & $0.0000{\pm}0.0000$ & $0.0000{\pm}0.0000$ & $0.0000{\pm}0.0000$ & $0.0000{\pm}0.0000$ & $0.9200{\pm}0.0980$ & None & 0.9200 \\
STTC & ISCA & 92 & $0.1151{\pm}0.0125$ & $0.1505{\pm}0.0100$ & $0.3704{\pm}0.0045$ & $0.2930{\pm}0.0115$ & $0.3065{\pm}0.0111$ & $0.3581{\pm}0.0097$ & $0.1821{\pm}0.0097$ & $0.8370{\pm}0.0059$ & TS2Vec & 0.4666 \\
STTC & ISCI & 40 & $0.0882{\pm}0.0110$ & $0.1667{\pm}0.0100$ & $0.4190{\pm}0.0176$ & $0.2407{\pm}0.0107$ & $0.3377{\pm}0.0102$ & $0.3433{\pm}0.0169$ & $0.1483{\pm}0.0381$ & $0.7500{\pm}0.0164$ & TS2Vec & 0.3310 \\
STTC & ISC\_ & 127 & $0.3049{\pm}0.0098$ & $0.4843{\pm}0.0055$ & $0.6806{\pm}0.0117$ & $0.5579{\pm}0.0100$ & $0.5570{\pm}0.0033$ & $0.6139{\pm}0.0074$ & $0.4316{\pm}0.0019$ & $0.9449{\pm}0.0058$ & TS2Vec & 0.2643 \\
STTC & NST\_ & 77 & $0.0727{\pm}0.0118$ & $0.0449{\pm}0.0077$ & $0.2660{\pm}0.0151$ & $0.1707{\pm}0.0156$ & $0.2929{\pm}0.0109$ & $0.1935{\pm}0.0114$ & $0.0746{\pm}0.0192$ & $0.8442{\pm}0.0058$ & UniTS & 0.5513 \\
STTC & STTC & 226 & $0.2615{\pm}0.0111$ & $0.3978{\pm}0.0056$ & $0.4794{\pm}0.0116$ & $0.4625{\pm}0.0105$ & $0.4510{\pm}0.0037$ & $0.4325{\pm}0.0105$ & $0.3399{\pm}0.0067$ & $0.7920{\pm}0.0065$ & TS2Vec & 0.3126 \\
CD & \_AVB & 83 & $0.0796{\pm}0.0114$ & $0.1606{\pm}0.0153$ & $0.4662{\pm}0.0145$ & $0.2364{\pm}0.0148$ & $0.4701{\pm}0.0127$ & $0.1648{\pm}0.0108$ & $0.1397{\pm}0.0066$ & $0.6265{\pm}0.0100$ & UniTS & 0.1564 \\
CD & CLBBB & 54 & $0.7172{\pm}0.0059$ & $0.5854{\pm}0.0136$ & $0.8947{\pm}0.0141$ & $0.8205{\pm}0.0119$ & $0.8430{\pm}0.0137$ & $0.8785{\pm}0.0147$ & $0.6869{\pm}0.0052$ & $0.9630{\pm}0.0137$ & TS2Vec & 0.0683 \\
CD & CRBBB & 55 & $0.6541{\pm}0.0121$ & $0.7321{\pm}0.0058$ & $0.8525{\pm}0.0140$ & $0.7737{\pm}0.0080$ & $0.8293{\pm}0.0103$ & $0.7597{\pm}0.0133$ & $0.7369{\pm}0.0080$ & $0.9091{\pm}0.0112$ & TS2Vec & 0.0566 \\
CD & ILBBB & 7 & $0.0000{\pm}0.0000$ & $0.0000{\pm}0.0000$ & $0.1250{\pm}0.0169$ & $0.0000{\pm}0.0000$ & $0.3750{\pm}0.0203$ & $0.2857{\pm}0.0229$ & $0.0000{\pm}0.0000$ & $0.8571{\pm}0.0199$ & UniTS & 0.4821 \\
CD & IRBBB & 112 & $0.5201{\pm}0.0125$ & $0.3486{\pm}0.0138$ & $0.5687{\pm}0.0115$ & $0.5918{\pm}0.0044$ & $0.5950{\pm}0.0120$ & $0.5077{\pm}0.0136$ & $0.3380{\pm}0.0087$ & $0.8036{\pm}0.0128$ & UniTS & 0.2086 \\
CD & IVCD & 79 & $0.1387{\pm}0.0115$ & $0.1101{\pm}0.0145$ & $0.2420{\pm}0.0110$ & $0.1299{\pm}0.0169$ & $0.1769{\pm}0.0064$ & $0.1808{\pm}0.0072$ & $0.1933{\pm}0.0134$ & $0.5190{\pm}0.0081$ & TS2Vec & 0.2770 \\
CD & LAFB/LPFB & 181 & $0.6511{\pm}0.0026$ & $0.5646{\pm}0.0067$ & $0.7420{\pm}0.0031$ & $0.6530{\pm}0.0091$ & $0.6536{\pm}0.0047$ & $0.7019{\pm}0.0038$ & $0.5206{\pm}0.0013$ & $0.9282{\pm}0.0041$ & TS2Vec & 0.1862 \\
CD & WPW & 8 & $0.1333{\pm}0.0233$ & $0.0000{\pm}0.0000$ & $0.2000{\pm}0.0250$ & $0.0000{\pm}0.0000$ & $0.4000{\pm}0.0179$ & $0.2000{\pm}0.0177$ & $0.0000{\pm}0.0000$ & $0.7500{\pm}0.0149$ & UniTS & 0.3500 \\
HYP & LAO/LAE & 43 & $0.0155{\pm}0.0161$ & $0.0000{\pm}0.0000$ & $0.2029{\pm}0.0117$ & $0.0213{\pm}0.0112$ & $0.1644{\pm}0.0120$ & $0.0959{\pm}0.0119$ & $0.0085{\pm}0.0148$ & $0.2791{\pm}0.0076$ & TS2Vec & 0.0762 \\
HYP & LVH & 213 & $0.3461{\pm}0.0109$ & $0.4935{\pm}0.0061$ & $0.6748{\pm}0.0034$ & $0.5806{\pm}0.0067$ & $0.4799{\pm}0.0112$ & $0.6497{\pm}0.0064$ & $0.4904{\pm}0.0024$ & $0.7559{\pm}0.0028$ & TS2Vec & 0.0811 \\
HYP & RAO/RAE & 10 & $0.0000{\pm}0.0000$ & $0.0000{\pm}0.0000$ & $0.2381{\pm}0.0194$ & $0.0000{\pm}0.0000$ & $0.2105{\pm}0.0219$ & $0.2222{\pm}0.0264$ & $0.1465{\pm}0.0063$ & $0.3000{\pm}0.0165$ & TS2Vec & 0.0619 \\
HYP & RVH & 12 & $0.1875{\pm}0.0175$ & $0.2295{\pm}0.0177$ & $0.2500{\pm}0.0168$ & $0.1702{\pm}0.0130$ & $0.1818{\pm}0.0076$ & $0.3226{\pm}0.0171$ & $0.3531{\pm}0.0091$ & $0.2500{\pm}0.0178$ & Argos & -0.1031 \\
HYP & SEHYP & 3 & $0.0000{\pm}0.0000$ & $0.0000{\pm}0.0000$ & $0.5714{\pm}0.0000$ & $0.0000{\pm}0.0000$ & $0.0000{\pm}0.0000$ & $0.0000{\pm}0.0000$ & $0.5000{\pm}0.0000$ & $0.3333{\pm}0.0242$ & TS2Vec & -0.2381 \\
\bottomrule
\end{tabular}}
\end{table*}

\begin{table*}[t]
\centering
\caption{Georgia label F1 results reported as mean~$\pm$~standard deviation over five runs. Delta is computed from mean values.}
\label{tab:app_georgia_label_latest}
\scriptsize
\setlength{\tabcolsep}{4pt}
\resizebox{0.94\textwidth}{!}{%
\begin{tabular}{lrrrrrrrrrlr}
\toprule
Label & Support & PatchTST & TimesNet & TS2Vec & xLSTM & UniTS & MERL & Argos & \sysname & Best baseline & Delta \\
\midrule
NSR & 192 & $0.6069{\pm}0.0055$ & $0.6129{\pm}0.0083$ & $0.7789{\pm}0.0021$ & $0.8413{\pm}0.0062$ & $0.6720{\pm}0.0036$ & $0.8751{\pm}0.0065$ & $0.6662{\pm}0.0060$ & $0.9514{\pm}0.0108$ & MERL & 0.0763 \\
AF & 70 & $0.2150{\pm}0.0164$ & $0.2687{\pm}0.0121$ & $0.3864{\pm}0.0137$ & $0.5306{\pm}0.0092$ & $0.2295{\pm}0.0075$ & $0.6854{\pm}0.0101$ & $0.2622{\pm}0.0114$ & $0.7761{\pm}0.0096$ & MERL & 0.0907 \\
IAVB & 61 & $0.2426{\pm}0.0109$ & $0.3751{\pm}0.0110$ & $0.3110{\pm}0.0098$ & $0.3504{\pm}0.0142$ & $0.4047{\pm}0.0101$ & $0.7968{\pm}0.0151$ & $0.3090{\pm}0.0253$ & $0.8387{\pm}0.0134$ & MERL & 0.0419 \\
LBBB & 18 & $0.7701{\pm}0.0143$ & $0.5092{\pm}0.0124$ & $0.6639{\pm}0.0184$ & $0.5947{\pm}0.0181$ & $0.6236{\pm}0.0139$ & $0.7833{\pm}0.0120$ & $0.6784{\pm}0.0103$ & $0.8571{\pm}0.0153$ & MERL & 0.0738 \\
RBBB & 41 & $0.7490{\pm}0.0123$ & $0.6173{\pm}0.0109$ & $0.6607{\pm}0.0101$ & $0.7738{\pm}0.0113$ & $0.5660{\pm}0.0184$ & $0.7981{\pm}0.0126$ & $0.7751{\pm}0.0097$ & $0.8395{\pm}0.0192$ & MERL & 0.0414 \\
SB & 146 & $0.6079{\pm}0.0049$ & $0.7354{\pm}0.0057$ & $0.7511{\pm}0.0079$ & $0.8203{\pm}0.0096$ & $0.6467{\pm}0.0067$ & $0.8187{\pm}0.0090$ & $0.6557{\pm}0.0000$ & $0.9667{\pm}0.0070$ & xLSTM & 0.1464 \\
STach & 137 & $0.7105{\pm}0.0090$ & $0.7866{\pm}0.0096$ & $0.7991{\pm}0.0114$ & $0.8676{\pm}0.0097$ & $0.7152{\pm}0.0097$ & $0.9333{\pm}0.0033$ & $0.7912{\pm}0.0050$ & $0.9485{\pm}0.0053$ & MERL & 0.0152 \\
\bottomrule
\end{tabular}}
\end{table*}

\begin{table*}[t]
\centering
\caption{CPSC2018 label F1 results reported as mean~$\pm$~standard deviation over five runs. Delta is computed from mean values.}
\label{tab:app_cpsc_label_latest}
\scriptsize
\setlength{\tabcolsep}{4pt}
\resizebox{0.94\textwidth}{!}{%
\begin{tabular}{lrrrrrrrrrlr}
\toprule
Label & Support & PatchTST & TimesNet & TS2Vec & xLSTM & UniTS & MERL & Argos & \sysname & Best baseline & Delta \\
\midrule
NSR & 184 & $0.3419{\pm}0.0112$ & $0.4109{\pm}0.0021$ & $0.5596{\pm}0.0080$ & $0.5662{\pm}0.0082$ & $0.4963{\pm}0.0032$ & $0.6779{\pm}0.0028$ & $0.5431{\pm}0.0084$ & $0.9112{\pm}0.0046$ & MERL & 0.2333 \\
AF & 237 & $0.4841{\pm}0.0104$ & $0.4171{\pm}0.0078$ & $0.5923{\pm}0.0032$ & $0.7869{\pm}0.0093$ & $0.6715{\pm}0.0081$ & $0.8718{\pm}0.0066$ & $0.5636{\pm}0.0147$ & $0.9494{\pm}0.0021$ & MERL & 0.0776 \\
IAVB & 144 & $0.2308{\pm}0.0078$ & $0.2115{\pm}0.0089$ & $0.2624{\pm}0.0028$ & $0.4075{\pm}0.0059$ & $0.4304{\pm}0.0112$ & $0.8291{\pm}0.0060$ & $0.3232{\pm}0.0385$ & $0.8529{\pm}0.0068$ & MERL & 0.0238 \\
LBBB & 372 & $0.7500{\pm}0.0071$ & $0.7333{\pm}0.0044$ & $0.7387{\pm}0.0068$ & $0.7636{\pm}0.0075$ & $0.8431{\pm}0.0047$ & $0.8723{\pm}0.0100$ & $0.8629{\pm}0.0050$ & $0.9667{\pm}0.0106$ & MERL & 0.0944 \\
RBBB & 45 & $0.8152{\pm}0.0164$ & $0.7658{\pm}0.0140$ & $0.8254{\pm}0.0144$ & $0.8667{\pm}0.0132$ & $0.8184{\pm}0.0126$ & $0.9084{\pm}0.0127$ & $0.7964{\pm}0.0111$ & $0.5556{\pm}0.0206$ & MERL & -0.3528 \\
PAC & 121 & $0.1951{\pm}0.0097$ & $0.1860{\pm}0.0109$ & $0.0448{\pm}0.0083$ & $0.2633{\pm}0.0051$ & $0.1709{\pm}0.0109$ & $0.3333{\pm}0.0067$ & $0.1635{\pm}0.0071$ & $0.6593{\pm}0.0093$ & MERL & 0.3260 \\
PVC & 142 & $0.5415{\pm}0.0104$ & $0.4120{\pm}0.0035$ & $0.6731{\pm}0.0072$ & $0.7101{\pm}0.0059$ & $0.6190{\pm}0.0101$ & $0.5414{\pm}0.0090$ & $0.6815{\pm}0.0099$ & $0.8021{\pm}0.0066$ & xLSTM & 0.0920 \\
STD & 43 & $0.3608{\pm}0.0112$ & $0.3508{\pm}0.0176$ & $0.5030{\pm}0.0188$ & $0.6114{\pm}0.0159$ & $0.5265{\pm}0.0108$ & $0.6907{\pm}0.0132$ & $0.5601{\pm}0.0115$ & $0.7737{\pm}0.0067$ & MERL & 0.0830 \\
STE & 180 & $0.2182{\pm}0.0106$ & $0.1923{\pm}0.0030$ & $0.2424{\pm}0.0086$ & $0.2222{\pm}0.0108$ & $0.3469{\pm}0.0069$ & $0.5128{\pm}0.0075$ & $0.3042{\pm}0.0211$ & $0.7534{\pm}0.0039$ & MERL & 0.2406 \\
\bottomrule
\end{tabular}}
\end{table*}

\begin{table*}[t]
\centering
\caption{PTB-XL diagnostic subclass AUC results reported as mean~$\pm$~standard deviation over five runs. Delta is computed from mean values.}
\label{tab:ptbxl-subclass-auc}
\scriptsize
\setlength{\tabcolsep}{3pt}
\resizebox{\textwidth}{!}{%
\begin{tabular}{llrrrrrrrrrlr}
\toprule
Superclass & Subclass & Support & PatchTST & TimesNet & TS2Vec & xLSTM & UniTS & MERL & Argos & \sysname & Best baseline & Delta \\
\midrule
NORM & NORM & 957 & $0.8632{\pm}0.0045$ & $0.8865{\pm}0.0036$ & $0.9055{\pm}0.0048$ & $0.9328{\pm}0.0034$ & $0.9225{\pm}0.0017$ & $0.9308{\pm}0.0019$ & $0.8777{\pm}0.0110$ & $0.9540{\pm}0.0017$ & xLSTM & 0.0212 \\
MI & AMI & 307 & $0.8865{\pm}0.0043$ & $0.8630{\pm}0.0033$ & $0.8617{\pm}0.0018$ & $0.9495{\pm}0.0044$ & $0.9431{\pm}0.0047$ & $0.9392{\pm}0.0021$ & $0.8659{\pm}0.0086$ & $0.9591{\pm}0.0015$ & xLSTM & 0.0096 \\
MI & IMI & 330 & $0.8567{\pm}0.0015$ & $0.8354{\pm}0.0052$ & $0.8452{\pm}0.0050$ & $0.9188{\pm}0.0022$ & $0.9074{\pm}0.0043$ & $0.9245{\pm}0.0053$ & $0.8272{\pm}0.0067$ & $0.9337{\pm}0.0023$ & MERL & 0.0092 \\
MI & LMI & 20 & $0.7278{\pm}0.0022$ & $0.7192{\pm}0.0046$ & $0.7677{\pm}0.0034$ & $0.8834{\pm}0.0054$ & $0.8812{\pm}0.0014$ & $0.8809{\pm}0.0040$ & $0.7975{\pm}0.0214$ & $0.9191{\pm}0.0017$ & xLSTM & 0.0357 \\
MI & PMI & 2 & $0.6872{\pm}0.0037$ & $0.6596{\pm}0.0051$ & $0.7783{\pm}0.0021$ & $0.6541{\pm}0.0063$ & $0.9137{\pm}0.0034$ & $0.8579{\pm}0.0029$ & $0.7789{\pm}0.1101$ & $0.9263{\pm}0.0026$ & UniTS & 0.0126 \\
STTC & ISCA & 92 & $0.7714{\pm}0.0023$ & $0.8387{\pm}0.0029$ & $0.8269{\pm}0.0014$ & $0.8905{\pm}0.0026$ & $0.9012{\pm}0.0031$ & $0.9052{\pm}0.0018$ & $0.7966{\pm}0.0048$ & $0.9581{\pm}0.0012$ & MERL & 0.0529 \\
STTC & ISCI & 40 & $0.7484{\pm}0.0055$ & $0.7789{\pm}0.0039$ & $0.8499{\pm}0.0016$ & $0.8772{\pm}0.0040$ & $0.8971{\pm}0.0019$ & $0.8820{\pm}0.0012$ & $0.8050{\pm}0.0151$ & $0.9205{\pm}0.0012$ & UniTS & 0.0234 \\
STTC & ISC\_ & 127 & $0.8709{\pm}0.0010$ & $0.9196{\pm}0.0029$ & $0.8639{\pm}0.0010$ & $0.9587{\pm}0.0050$ & $0.9459{\pm}0.0012$ & $0.9465{\pm}0.0033$ & $0.9135{\pm}0.0119$ & $0.9587{\pm}0.0047$ & xLSTM & 0.0000 \\
STTC & NST\_ & 77 & $0.6682{\pm}0.0017$ & $0.7828{\pm}0.0033$ & $0.7473{\pm}0.0069$ & $0.7861{\pm}0.0023$ & $0.8361{\pm}0.0033$ & $0.8297{\pm}0.0052$ & $0.7009{\pm}0.0055$ & $0.9124{\pm}0.0026$ & UniTS & 0.0763 \\
STTC & STTC & 226 & $0.7335{\pm}0.0052$ & $0.8141{\pm}0.0047$ & $0.7963{\pm}0.0050$ & $0.8821{\pm}0.0053$ & $0.8672{\pm}0.0028$ & $0.8709{\pm}0.0020$ & $0.8236{\pm}0.0006$ & $0.9454{\pm}0.0049$ & xLSTM & 0.0633 \\
CD & \_AVB & 83 & $0.6786{\pm}0.0051$ & $0.7115{\pm}0.0043$ & $0.8655{\pm}0.0048$ & $0.7790{\pm}0.0047$ & $0.9495{\pm}0.0023$ & $0.7694{\pm}0.0059$ & $0.6962{\pm}0.0065$ & $0.9515{\pm}0.0012$ & UniTS & 0.0020 \\
CD & CLBBB & 54 & $0.9952{\pm}0.0037$ & $0.9884{\pm}0.0012$ & $0.8971{\pm}0.0041$ & $0.9964{\pm}0.0022$ & $0.9983{\pm}0.0025$ & $0.9983{\pm}0.0025$ & $0.9970{\pm}0.0002$ & $0.9983{\pm}0.0019$ & UniTS/MERL & 0.0000 \\
CD & CRBBB & 55 & $0.9952{\pm}0.0027$ & $0.9914{\pm}0.0030$ & $0.8974{\pm}0.0058$ & $0.9972{\pm}0.0033$ & $0.9970{\pm}0.0010$ & $0.9965{\pm}0.0017$ & $0.9933{\pm}0.0004$ & $0.9718{\pm}0.0041$ & xLSTM & -0.0254 \\
CD & ILBBB & 7 & $0.8198{\pm}0.0021$ & $0.8695{\pm}0.0015$ & $0.8268{\pm}0.0023$ & $0.8869{\pm}0.0046$ & $0.9110{\pm}0.0045$ & $0.8567{\pm}0.0047$ & $0.9117{\pm}0.0333$ & $0.8933{\pm}0.0015$ & Argos & -0.0184 \\
CD & IRBBB & 112 & $0.9273{\pm}0.0057$ & $0.8483{\pm}0.0051$ & $0.8520{\pm}0.0051$ & $0.9515{\pm}0.0039$ & $0.9588{\pm}0.0042$ & $0.9484{\pm}0.0018$ & $0.8683{\pm}0.0041$ & $0.9704{\pm}0.0032$ & UniTS & 0.0116 \\
CD & IVCD & 79 & $0.6496{\pm}0.0032$ & $0.6495{\pm}0.0044$ & $0.6281{\pm}0.0061$ & $0.6487{\pm}0.0054$ & $0.7018{\pm}0.0021$ & $0.7401{\pm}0.0042$ & $0.6627{\pm}0.0043$ & $0.8828{\pm}0.0046$ & MERL & 0.1427 \\
CD & LAFB/LPFB & 181 & $0.9513{\pm}0.0026$ & $0.9432{\pm}0.0048$ & $0.8640{\pm}0.0042$ & $0.9519{\pm}0.0017$ & $0.9665{\pm}0.0015$ & $0.9455{\pm}0.0037$ & $0.9609{\pm}0.0231$ & $0.9516{\pm}0.0040$ & UniTS & -0.0149 \\
CD & WPW & 8 & $0.7547{\pm}0.0056$ & $0.5929{\pm}0.0056$ & $0.7537{\pm}0.0044$ & $0.7831{\pm}0.0039$ & $0.9181{\pm}0.0033$ & $0.8386{\pm}0.0038$ & $0.7831{\pm}0.0641$ & $0.9672{\pm}0.0015$ & UniTS & 0.0491 \\
HYP & LAO/LAE & 43 & $0.6961{\pm}0.0055$ & $0.7273{\pm}0.0026$ & $0.7059{\pm}0.0045$ & $0.7258{\pm}0.0043$ & $0.8092{\pm}0.0031$ & $0.7653{\pm}0.0054$ & $0.7043{\pm}0.0123$ & $0.8247{\pm}0.0021$ & UniTS & 0.0155 \\
HYP & LVH & 213 & $0.7981{\pm}0.0054$ & $0.8881{\pm}0.0033$ & $0.8473{\pm}0.0031$ & $0.9254{\pm}0.0024$ & $0.8778{\pm}0.0039$ & $0.9417{\pm}0.0037$ & $0.9318{\pm}0.0009$ & $0.9419{\pm}0.0028$ & MERL & 0.0002 \\
HYP & RAO/RAE & 10 & $0.7090{\pm}0.0042$ & $0.6974{\pm}0.0027$ & $0.8700{\pm}0.0016$ & $0.7632{\pm}0.0016$ & $0.8980{\pm}0.0029$ & $0.9018{\pm}0.0037$ & $0.8818{\pm}0.0426$ & $0.9381{\pm}0.0055$ & MERL & 0.0363 \\
HYP & RVH & 12 & $0.9264{\pm}0.0015$ & $0.8711{\pm}0.0051$ & $0.8516{\pm}0.0049$ & $0.8602{\pm}0.0031$ & $0.9689{\pm}0.0013$ & $0.9195{\pm}0.0053$ & $0.9143{\pm}0.0084$ & $0.9885{\pm}0.0031$ & UniTS & 0.0196 \\
HYP & SEHYP & 3 & $0.8031{\pm}0.0048$ & $0.7017{\pm}0.0058$ & $0.8890{\pm}0.0054$ & $0.8083{\pm}0.0033$ & $0.9340{\pm}0.0043$ & $0.9238{\pm}0.0050$ & $0.9961{\pm}0.0013$ & $0.9613{\pm}0.0057$ & Argos & -0.0348 \\
\bottomrule
\end{tabular}}
\end{table*}

\begin{table*}[t]
\centering
\caption{Georgia label AUC results reported as mean~$\pm$~standard deviation over five runs. Delta is computed from mean values.}
\label{tab:georgia-label-auc}
\scriptsize
\setlength{\tabcolsep}{4pt}
\resizebox{0.94\textwidth}{!}{%
\begin{tabular}{lrrrrrrrrrlr}
\toprule
Label & Support & PatchTST & TimesNet & TS2Vec & xLSTM & UniTS & MERL & Argos & \sysname & Best baseline & Delta \\
\midrule
NSR & 192 & $0.7716{\pm}0.0038$ & $0.8007{\pm}0.0022$ & $0.9247{\pm}0.0030$ & $0.9702{\pm}0.0026$ & $0.7277{\pm}0.0058$ & $0.9756{\pm}0.0010$ & $0.8427{\pm}0.0012$ & $0.9934{\pm}0.0012$ & MERL & 0.0178 \\
AF & 70 & $0.6760{\pm}0.0057$ & $0.6908{\pm}0.0062$ & $0.8646{\pm}0.0050$ & $0.8671{\pm}0.0050$ & $0.5445{\pm}0.0016$ & $0.9331{\pm}0.0026$ & $0.7840{\pm}0.0077$ & $0.9455{\pm}0.0011$ & MERL & 0.0124 \\
IAVB & 61 & $0.7235{\pm}0.0021$ & $0.7928{\pm}0.0069$ & $0.7647{\pm}0.0019$ & $0.7717{\pm}0.0061$ & $0.4937{\pm}0.0051$ & $0.9773{\pm}0.0024$ & $0.7765{\pm}0.0060$ & $0.9775{\pm}0.0035$ & MERL & 0.0002 \\
LBBB & 18 & $0.9920{\pm}0.0025$ & $0.8912{\pm}0.0029$ & $0.9890{\pm}0.0021$ & $0.9341{\pm}0.0035$ & $0.8382{\pm}0.0046$ & $0.9637{\pm}0.0040$ & $0.9746{\pm}0.0016$ & $0.9988{\pm}0.0014$ & PatchTST & 0.0068 \\
RBBB & 41 & $0.9447{\pm}0.0018$ & $0.9199{\pm}0.0040$ & $0.9625{\pm}0.0028$ & $0.9643{\pm}0.0051$ & $0.8414{\pm}0.0027$ & $0.9607{\pm}0.0047$ & $0.9440{\pm}0.0032$ & $0.9592{\pm}0.0013$ & xLSTM & -0.0051 \\
SB & 146 & $0.8946{\pm}0.0038$ & $0.9194{\pm}0.0027$ & $0.9547{\pm}0.0013$ & $0.9745{\pm}0.0027$ & $0.7131{\pm}0.0039$ & $0.9843{\pm}0.0012$ & $0.9164{\pm}0.0016$ & $0.9981{\pm}0.0017$ & MERL & 0.0138 \\
STach & 137 & $0.9189{\pm}0.0025$ & $0.9503{\pm}0.0038$ & $0.9536{\pm}0.0045$ & $0.9814{\pm}0.0010$ & $0.8460{\pm}0.0021$ & $0.9902{\pm}0.0046$ & $0.9558{\pm}0.0010$ & $0.9976{\pm}0.0028$ & MERL & 0.0074 \\
\bottomrule
\end{tabular}}
\end{table*}

\begin{table*}[t]
\centering
\caption{CPSC2018 label AUC results reported as mean~$\pm$~standard deviation over five runs. Delta is computed from mean values.}
\label{tab:cpsc2018-label-auc}
\scriptsize
\setlength{\tabcolsep}{4pt}
\resizebox{0.94\textwidth}{!}{%
\begin{tabular}{lrrrrrrrrrlr}
\toprule
Label & Support & PatchTST & TimesNet & TS2Vec & xLSTM & UniTS & MERL & Argos & \sysname & Best baseline & Delta \\
\midrule
NSR & 184 & $0.7533{\pm}0.0053$ & $0.7617{\pm}0.0050$ & $0.8693{\pm}0.0026$ & $0.9368{\pm}0.0014$ & $0.8762{\pm}0.0046$ & $0.8190{\pm}0.0010$ & $0.8755{\pm}0.0021$ & $0.9880{\pm}0.0024$ & xLSTM & 0.0512 \\
AF & 237 & $0.7869{\pm}0.0065$ & $0.6933{\pm}0.0056$ & $0.8722{\pm}0.0019$ & $0.9673{\pm}0.0025$ & $0.9067{\pm}0.0035$ & $0.7953{\pm}0.0041$ & $0.8476{\pm}0.0020$ & $0.9944{\pm}0.0009$ & xLSTM & 0.0271 \\
IAVB & 144 & $0.6987{\pm}0.0049$ & $0.6304{\pm}0.0042$ & $0.7641{\pm}0.0013$ & $0.8924{\pm}0.0041$ & $0.8054{\pm}0.0022$ & $0.7467{\pm}0.0057$ & $0.7964{\pm}0.0011$ & $0.9825{\pm}0.0034$ & xLSTM & 0.0901 \\
LBBB & 372 & $0.9818{\pm}0.0042$ & $0.9190{\pm}0.0038$ & $0.9740{\pm}0.0045$ & $0.9790{\pm}0.0028$ & $0.9772{\pm}0.0022$ & $0.9615{\pm}0.0032$ & $0.9857{\pm}0.0020$ & $0.9995{\pm}0.0010$ & Argos & 0.0138 \\
RBBB & 45 & $0.9401{\pm}0.0046$ & $0.9045{\pm}0.0053$ & $0.9622{\pm}0.0024$ & $0.9846{\pm}0.0023$ & $0.9612{\pm}0.0025$ & $0.9050{\pm}0.0042$ & $0.9495{\pm}0.0006$ & $0.9721{\pm}0.0022$ & xLSTM & -0.0125 \\
PAC & 121 & $0.5698{\pm}0.0026$ & $0.6143{\pm}0.0046$ & $0.6511{\pm}0.0034$ & $0.7477{\pm}0.0023$ & $0.6848{\pm}0.0065$ & $0.6358{\pm}0.0054$ & $0.7374{\pm}0.0073$ & $0.9502{\pm}0.0039$ & xLSTM & 0.2025 \\
PVC & 142 & $0.8617{\pm}0.0052$ & $0.7092{\pm}0.0043$ & $0.8869{\pm}0.0043$ & $0.9025{\pm}0.0028$ & $0.8726{\pm}0.0018$ & $0.8068{\pm}0.0032$ & $0.8981{\pm}0.0029$ & $0.9611{\pm}0.0045$ & xLSTM & 0.0586 \\
STD & 43 & $0.7191{\pm}0.0042$ & $0.6771{\pm}0.0019$ & $0.8293{\pm}0.0032$ & $0.8928{\pm}0.0018$ & $0.8534{\pm}0.0033$ & $0.7342{\pm}0.0035$ & $0.8651{\pm}0.0013$ & $0.9665{\pm}0.0030$ & xLSTM & 0.0737 \\
STE & 180 & $0.7445{\pm}0.0042$ & $0.6469{\pm}0.0026$ & $0.7420{\pm}0.0029$ & $0.8114{\pm}0.0037$ & $0.8638{\pm}0.0041$ & $0.8384{\pm}0.0038$ & $0.8819{\pm}0.0076$ & $0.9575{\pm}0.0016$ & Argos & 0.0756 \\
\bottomrule
\end{tabular}}
\end{table*}

The label-level evidence clarifies the aggregate result.
\sysname{} is strongest on labels for which failures analyzed
through Evidence-Grounded Failure Review can be grounded in
reproducible morphology, rhythm, or cross-lead measurements. EGFR
does not infer the cause of an error from the label mismatch alone;
it jointly considers raw waveforms, reference-backed measurements,
model outputs, and comparator cases.

The complementary metrics also reveal different failure modes.
PTB-XL RVH trails the strongest baseline in F1 but exceeds it in AUC,
suggesting that its remaining weakness may be related to the operating
point, threshold selection, or calibration rather than a complete loss
of discriminative ranking. Conversely, the AUC regressions on PTB-XL
CRBBB, ILBBB, LAFB/LPFB, and SEHYP, Georgia RBBB, and CPSC2018 RBBB
show that \sysname{} is not uniformly dominant in ranking quality.

Several of these labels have very low support, so their estimates
should be interpreted cautiously. We report all labels and both
metrics to make these exceptions explicit rather than excluding
unfavorable cases post hoc.

\subsection{Failure Modes and Negative-Label Analysis}
\label{app:failure_modes_latest}

The negative label-level deltas identify where the current workflow
still leaves label-specific errors, rather than where the overall
method fails. Across the three datasets, the principal F1 exceptions
are CPSC2018 RBBB and the low-support PTB-XL hypertrophy subclasses
RVH and SEHYP. These labels are harder than the high-gain cases
because their evidence is either sparsely represented, heterogeneous
across cases, or less directly captured by the retained measurement
and revision path.

\fakeparagraph{CPSC2018 RBBB}
RBBB is the clearest negative case in the label-level comparison.
\sysname{} reaches an F1 of 0.5556, whereas MERL reaches 0.9084 on
the same evaluation split. This exception does not overturn the
aggregate CPSC2018 gain, where \sysname{} improves macro F1 from
0.6931 to 0.8027 and improves eight of the nine evaluated labels,
but it shows that the learned decision rule is not uniformly reliable
across all labels.

Reliable bundle-branch-block characterization depends on QRS
duration, terminal forces, and lead-specific morphology in V1--V2
and lateral leads. Although \sysname{} improves several arrhythmia
and ST-segment labels, the RBBB drop appears in both F1 and AUC.
This suggests a label-specific ranking and calibration problem, not
only an unfavorable threshold.

The ablation results show the same local weakness. For RBBB, each
ablated variant obtains higher F1 than the full system, even though
the full system gives the best aggregate CPSC2018 result. This pattern
suggests a trade-off between aggregate optimization and some
label-wise decision boundaries.

RBBB should therefore be treated as a target for more specialized
conduction-block measurement functions, stronger label-wise
regression checks, and separate failure-selection quotas within EGFR.

\fakeparagraph{PTB-XL RVH and SEHYP}
The PTB-XL negative labels differ from CPSC2018 RBBB because their
results are dominated by support constraints. RVH has 12 test cases
and SEHYP has 3 test cases in the diagnostic-subclass evaluation.
At this scale, changing the prediction of a single case can cause a
large F1 movement. These rows are therefore useful as audit signals
but should not be interpreted as stable estimates of diagnostic
superiority.

The negative deltas do not imply that the framework cannot model
hypertrophy. The related LVH label has a support of 213 and improves
from the strongest-baseline F1 of 0.6748 to 0.7559.

The case-level evidence suggests a more specific limitation.
Hypertrophy subtyping requires the joint interpretation of amplitude
criteria, axis deviation, chamber-specific patterns, and secondary
repolarization changes. The current compiled measurement set
captures voltage and cross-lead evidence sufficiently to improve
LVH, but the rare right-ventricular and severe-hypertrophy
subclasses provide too few informative failures for stable
subtype-specific revision. This is primarily a data-efficiency
boundary of the present evaluation rather than direct evidence that
the self-iteration mechanism fails.

\fakeparagraph{Effect of imbalance}
Failure-case selection within EGFR could in principle overemphasize
frequent error patterns. RecursiveECG mitigates this risk by
selecting false positives, false negatives, low-margin predictions,
calibration errors, and label-specific failures rather than simply
reviewing the most numerous aggregate errors.

The label-level results are consistent with this design. Large gains
occur on several minority- or moderate-support labels, including
PTB-XL NST\_, ISCA, and WPW and CPSC2018 PAC and STE. The remaining
negative labels show that imbalance is not fully resolved, but the
observed failure pattern is localized rather than systemic.

These observations refine the claim of \sysname{}. The method is
most effective when failed cases expose reproducible morphology,
rhythm, or cross-lead evidence that can be compiled into
measurements and replayed during review. When a label has extremely
low support or depends on unstable, baseline-sensitive morphology,
the workflow records the evidence boundary rather than converting
an inconclusive case into a speculative rule. Additional
label-specific measurement functions or failure-selection quotas
may nevertheless be required. This behavior is consistent with the
intended role of \sysname{} as an auditable model-development
workflow rather than a claim of uniform per-label dominance.

\section{Mechanism Ablations and Iterative Refinement Process}
\label{app:mechanism_process}

\subsection{Mechanism Ablations}
\label{app:ablation_latest}

The ablation study reports the full mechanism-sensitivity
evaluation used in the main text. \sysname is the complete
workflow used for the main results and serves as the reference
configuration. We evaluate four controlled workflow variants that
remove one design choice at a time. The comparisons hold the candidate
budget, iteration budget, data splits, evaluation protocol, and final
predictor interface fixed. The two
core evidence-grounding mechanisms are Criteria-to-Measurement
Compilation (CMC) and Evidence-Grounded Failure Review (EGFR).

\begin{itemize}

\item \textbf{\emph{w/o} EGFR} removes informative failure-case
selection and the structured Evidence-Grounded Failure Review
procedure. The agent retains access to the validated measurement
functions produced by CMC, but it cannot jointly inspect selected
failures, comparator cases, raw waveforms, reference-backed
measurements, and model behavior to formulate revision hypotheses.
Candidate revisions must therefore be selected primarily from
aggregate validation metrics and candidate-level evaluation
summaries. This variant isolates the contribution of
failure-evidence-grounded revision.

\item \textbf{\emph{w/o} CMC} removes Criteria-to-Measurement
Compilation, including the compilation and validation of curated
ECG criteria as deterministic measurement functions. Domain
knowledge remains accessible through a direct reference-reading
interface, but EGFR receives no reproducible, reference-backed
measurement records emitted from individual ECGs. It can still
inspect raw waveforms, model outputs, and comparator cases. This
variant isolates the contribution of operationalizing textual
clinical criteria as validated deterministic measurements.

\item \textbf{\emph{w/o} Measurement Execution} retains the
measurement specifications, validated function implementations,
computation rules, and evidence back-pointers produced by CMC,
but prevents these functions from being executed on individual
review cases. EGFR therefore has access to the static measurement
definitions but not to their emitted case-level measurement
records. It must reason from raw waveforms, model outputs,
comparator cases, and static rule descriptions. This variant
distinguishes the availability of compiled measurement knowledge
from the use of executed measurement evidence during failure
review.

\item \textbf{\emph{Direct LLM Design}} replaces the governed workflow
with a single coding-agent session. It removes the explicit
problem-contract gate, CMC, independent candidate exploration,
structured EGFR, promotion ledger, and adaptive stopping
procedure. Candidate changes are not organized through the same
separation between evidence collection, revision formulation,
execution, re-evaluation, and promotion. This configuration serves
as a simplified workflow baseline and isolates the contribution of
the governed multi-stage organization.

\end{itemize}

Together, these five configurations provide the controlled
comparison used in the main text. The \emph{w/o EGFR} and
\emph{w/o CMC} variants isolate the two core evidence-grounding
mechanisms. The \emph{w/o Measurement Execution} variant separates
compiled-measurement availability from case-level measurement
usage. The \emph{Direct LLM Design} variant evaluates the contribution
of the governed workflow organization. Backbone sensitivity is reported
separately in the capability-gradient table below, while \sysname anchors
the mechanism comparison. The
rightmost column of
\tabref{tab:app_overall_ablation_latest} reports the Macro F1
difference between \sysname and each variant. Positive values
indicate degradation after ablation, whereas negative values
indicate that the ablated run exceeds the reference run on that
dataset.

\begin{table*}[t]
\centering
\caption{Overall mechanism-ablation results reported as mean~$\pm$~standard deviation over five runs. Delta is \sysname minus the
variant Macro F1 mean; positive values indicate degradation. Backbone
sensitivity is reported separately in the capability-gradient table.}
\label{tab:app_overall_ablation_latest}
\small
\begin{tabular}{llrrrrr}
\toprule
Dataset & Variant & Macro AUC & Macro F1 & Micro F1 &
Hamming Acc. & Delta Macro F1 \\
\midrule
PTB-XL & \sysname & $0.9373{\pm}0.0066$ & $0.7653{\pm}0.0088$ & $0.7930{\pm}0.0039$ & $0.8929{\pm}0.0177$ & 0.0000 \\
PTB-XL & w/o EGFR & $0.9149{\pm}0.0048$ & $0.7040{\pm}0.0093$ & $0.7546{\pm}0.0061$ & $0.8783{\pm}0.0084$ & 0.0613 \\
PTB-XL & w/o CMC & $0.9235{\pm}0.0037$ & $0.7354{\pm}0.0071$ & $0.7682{\pm}0.0054$ & $0.8789{\pm}0.0068$ & 0.0299 \\
PTB-XL & w/o Measurement Execution &
$0.9273{\pm}0.0042$ & $0.7430{\pm}0.0066$ & $0.7761{\pm}0.0047$ & $0.8892{\pm}0.0059$ & 0.0223 \\
PTB-XL & Direct LLM Design &
$0.9060{\pm}0.0055$ & $0.7047{\pm}0.0108$ & $0.7392{\pm}0.0070$ & $0.8505{\pm}0.0091$ & 0.0606 \\
Georgia & \sysname &
$0.9814{\pm}0.0052$ & $0.8826{\pm}0.0069$ & $0.9170{\pm}0.0043$ & $0.9735{\pm}0.0021$ & 0.0000 \\
Georgia & w/o EGFR &
$0.9788{\pm}0.0016$ & $0.8735{\pm}0.0057$ & $0.9220{\pm}0.0038$ & $0.9757{\pm}0.0025$ & 0.0091 \\
Georgia & w/o CMC &
$0.9765{\pm}0.0024$ & $0.8403{\pm}0.0082$ & $0.8951{\pm}0.0049$ & $0.9675{\pm}0.0031$ & 0.0423 \\
Georgia & w/o Measurement Execution &
$0.9808{\pm}0.0015$ & $0.8829{\pm}0.0064$ & $0.9148{\pm}0.0035$ & $0.9731{\pm}0.0022$ & -0.0003 \\
Georgia & Direct LLM Design &
$0.9223{\pm}0.0049$ & $0.6302{\pm}0.0116$ & $0.7918{\pm}0.0067$ & $0.9376{\pm}0.0055$ & 0.2524 \\
CPSC2018 & \sysname &
$0.9746{\pm}0.0036$ & $0.8027{\pm}0.0078$ & $0.8352{\pm}0.0056$ & $0.9613{\pm}0.0030$ & 0.0000 \\
CPSC2018 & w/o EGFR &
$0.9728{\pm}0.0020$ & $0.7616{\pm}0.0089$ & $0.7987{\pm}0.0063$ & $0.9444{\pm}0.0046$ & 0.0411 \\
CPSC2018 & w/o CMC &
$0.9735{\pm}0.0018$ & $0.7801{\pm}0.0073$ & $0.8105{\pm}0.0058$ & $0.9517{\pm}0.0039$ & 0.0226 \\
CPSC2018 & w/o Measurement Execution &
$0.9652{\pm}0.0031$ & $0.7928{\pm}0.0061$ & $0.8329{\pm}0.0045$ & $0.9609{\pm}0.0032$ & 0.0099 \\
CPSC2018 & Direct LLM Design &
$0.9663{\pm}0.0027$ & $0.7560{\pm}0.0096$ & $0.8182{\pm}0.0069$ & $0.9588{\pm}0.0036$ & 0.0467 \\
\bottomrule
\end{tabular}
\end{table*}

The dominant ablation effect differs across datasets. On PTB-XL,
removing EGFR reduces Macro F1 by 0.0613, while replacing the
governed workflow with Direct LLM reduces it by 0.0606. This
indicates that both structured failure review and workflow
organization contribute substantially in this setting. On Georgia,
the largest degradation is produced by Direct LLM Design (0.2524),
followed by the CMC ablation (0.0423), whereas disabling per-case
measurement execution is effectively neutral in this run. CPSC2018
shows lower aggregate sensitivity than Georgia but still degrades
under the workflow variants: Direct LLM Design reduces Macro F1 by 0.0467,
removing EGFR by 0.0411, removing CMC by
0.0226, and disabling per-case measurement execution by 0.0099. These
results support an aggregate mechanism claim rather than requiring
every workflow mechanism to improve every dataset or label. Backbone
sensitivity is summarized separately in
\tabref{tab:app_backbone_capability_gradient}.

\begin{table*}[t]
\centering
\caption{LLM-backbone sensitivity across three datasets. All metrics are reported as mean~$\pm$~standard deviation over five runs. The candidate budget and maximum iteration count are fixed at $k=3$ and 5, respectively.}
\label{tab:app_backbone_capability_gradient}
\small
\setlength{\tabcolsep}{4pt}
\begin{tabular}{ll lccrrrr}
\toprule
Dataset & Backbone & Capability tier & $k$ & Max iter. & Macro AUC & Macro F1 & Micro F1 & Hamming Acc. \\
\midrule
PTB-XL & DeepSeek-V4-Pro & Advanced reasoning & 3 & 5 & $0.9373{\pm}0.0066$ & $0.7653{\pm}0.0088$ & $0.7930{\pm}0.0039$ & $0.8929{\pm}0.0177$ \\
Georgia & DeepSeek-V4-Pro & Advanced reasoning & 3 & 5 & $0.9814{\pm}0.0052$ & $0.8826{\pm}0.0069$ & $0.9170{\pm}0.0043$ & $0.9735{\pm}0.0021$ \\
CPSC2018 & DeepSeek-V4-Pro & Advanced reasoning & 3 & 5 & $0.9746{\pm}0.0036$ & $0.8027{\pm}0.0078$ & $0.8352{\pm}0.0056$ & $0.9613{\pm}0.0030$ \\
\midrule
PTB-XL & Qwen3.5-27B & Lightweight reasoning & 3 & 5 & $0.9166{\pm}0.0187$ & $0.7338{\pm}0.0415$ & $0.7821{\pm}0.0326$ & $0.9118{\pm}0.0149$ \\
Georgia & Qwen3.5-27B & Lightweight reasoning & 3 & 5 & $0.9698{\pm}0.0124$ & $0.8021{\pm}0.0368$ & $0.8662{\pm}0.0297$ & $0.9587{\pm}0.0116$ \\
CPSC2018 & Qwen3.5-27B & Lightweight reasoning & 3 & 5 & $0.9437{\pm}0.0169$ & $0.6988{\pm}0.0442$ & $0.7204{\pm}0.0395$ & $0.9199{\pm}0.0173$ \\
\bottomrule
\end{tabular}
\end{table*}

The aggregate mechanism effects are not attributable to a single label.
PTB-XL is most consistently sensitive to removing EGFR, while Georgia
shows the clearest dependence on the governed workflow. CPSC2018 has a
more heterogeneous profile, with some labels neutral or improved under
individual ablations. The intended conclusion is robustness at the
aggregate and mechanism levels, rather than uniform per-label dominance
by every component.
\subsection{Search Width, Backbone, and Iterative Refinement Process}
\label{app:search_process_latest}

Search-width sensitivity indicates that $k=3$ provides the best
accuracy--cost operating point among the evaluated PTB-XL
settings. Increasing the search width to $k=5$ increases the number
of candidates that pass the execution and regression checks but does not
improve the final Macro F1,
whereas the single-candidate setting $k=1$ produces a weaker final
predictor.
The pass/fail counts in \tabref{tab:app_k_sensitivity_latest} summarize
candidate-level checks and should not be read as the number of promoted
revisions in the main iterative chain.

The resulting trajectory across the five refinement rounds is reported in
\tabref{tab:app_iteration_latest}. It separates the evidence or update
source used at each round from the final performance change, making clear
how the frozen predictor was selected rather than presenting only the final
endpoint.

\begin{table*}[t]
\centering
\caption{PTB-XL candidate-count sensitivity.}
\label{tab:app_k_sensitivity_latest}
\small
\setlength{\tabcolsep}{5pt}
\begin{tabularx}{0.92\textwidth}
{lrrrrrr>{\raggedright\arraybackslash}X}
\toprule
$k$ & Max iter. & AUC & Macro F1 & Micro F1 & Hamming Acc. &
Passed / failed checks & Note \\
\midrule
1 & 5 & 0.9231 & 0.7265 & 0.7638 & 0.8822 &
3 / 2 & greedy single-candidate update \\
3 & 5 & 0.9373 & 0.7653 & 0.7930 & 0.8929 &
5 / 10 & default setting \\
5 & 5 & 0.9325 & 0.7584 & 0.7869 & 0.8927 &
12 / 13 & wider search without a final Macro-F1 gain \\
\bottomrule
\end{tabularx}
\end{table*}

\begin{figure*}[t]
\centering
\includegraphics[width=0.70\textwidth]
{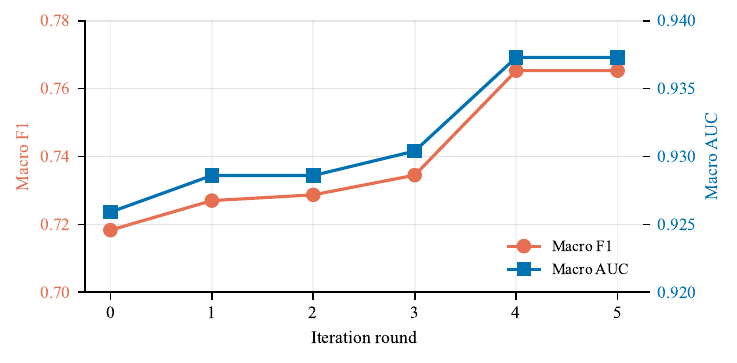}
\caption{PTB-XL iterative refinement trajectory.}
\Description{Line chart of Macro F1 and Macro AUC over the
initial configuration and five iterative refinement rounds.}
\label{fig:app_iteration_curve_latest}
\end{figure*}

\begin{table*}[t]
\centering
\caption{PTB-XL iterative refinement trajectory.}
\label{tab:app_iteration_latest}
\scriptsize
\setlength{\tabcolsep}{3pt}
\resizebox{\textwidth}{!}{%
\begin{tabular}{llrrrrr}
\toprule
Round & Main evidence or update source & Macro AUC & Macro F1 &
Micro F1 & Hamming Acc. & Gain over Round 0 \\
\midrule
0 & initial model and default threshold &
0.9259 & 0.7183 & 0.7641 & 0.8811 & 0.0000 \\

1 & informative validation failures and initial EGFR &
0.9286 & 0.7270 & 0.7723 & 0.8871 & 0.0087 \\

2 & reference-backed ECG measurements and label-wise error analysis &
0.9286 & 0.7287 & 0.7723 & 0.8871 & 0.0104 \\

3 & minority-label and false-negative failure review &
0.9304 & 0.7345 & 0.7700 & 0.8865 & 0.0162 \\

4 & regression-aware candidate evaluation &
0.9373 & 0.7653 & 0.7930 & 0.8929 & 0.0470 \\

5 & final frozen \sysname &
0.9373 & 0.7653 & 0.7930 & 0.8929 & 0.0470 \\

\bottomrule
\end{tabular}}
\end{table*}

The trajectory shows that the final predictor was not selected from
a single isolated trial. The largest improvement occurs after the
regression-aware revision in Round 4. This is consistent with the
method definition: EGFR formulates targeted revision hypotheses
from failure evidence, but a hypothesis does not become an update
solely because it is supported by an individual case. It must first
be implemented, executed under the fixed problem contract,
re-evaluated on the prescribed validation protocol, and accepted by
the regression and promotion gates before it enters the frozen
toolchain.

\section{Case Audits and Deployment}
\label{app:case_deployment}

\subsection{WPW Pseudo-Infarction Failure Audit}
\label{app:wpw_case_review}

\begin{table*}[t]
\centering
\caption{Complete retained audit trace for WPW case 461. The reference-backed measurements are computed from the supplied 12-lead ECG array around the reviewed QRS complex.}
\label{tab:app_wpw_case_review}
\small
\setlength{\tabcolsep}{5pt}
\begin{tabularx}{0.96\textwidth}{
    >{\raggedright\arraybackslash}p{0.24\textwidth}
    >{\raggedright\arraybackslash}X}
\toprule
Audit item & Evidence \\
\midrule

Sample &
Case 461, represented as a 12-lead ECG array with shape
$12 \times 1000$, sampled at 100 Hz and stored in mV. \\

Ground-truth condition &
WPW / ventricular pre-excitation. The case is reviewed because
WPW can produce pseudo-infarction morphology that resembles
myocardial infarction in individual leads. \\

Failure selected for Evidence-Grounded Failure Review &
The evaluation stage selected the record as an informative MI-like
error. Its local morphology contains Q-like initial negative
deflections, whereas the ground-truth condition is ventricular
pre-excitation. \\

Raw waveform evidence &
The reviewed record contains WPW-compatible pre-excitation evidence,
including a short PR interval, a slurred QRS onset consistent with a
delta wave, and a widened QRS complex. At the same time, several
leads show local pseudo-infarction morphology, which provides
plausible evidence for the initial MI-like prediction. \\

Reference-backed measurements &
The replayed measurements match the evidence shown in
\figref{fig:case_wpw_pseudo_mi}: the PR interval is below the
clinical short-PR threshold of 120 ms, the QRS duration is 130 ms,
and the frontal-plane axis is approximately $-35^{\circ}$, pointing
toward the left-superior/aVL sector. The QRS onset is slurred,
supporting a delta-wave interpretation rather than a purely
infarction-driven explanation. \\

Identified pipeline weakness &
The current predictor overweights lead-local pseudo-infarction
morphology and does not adequately evaluate whether the activation
pattern is consistent across limb and precordial leads. \\

Evidence-grounded interpretation &
The review links the apparent infarction cues to WPW physiology:
activation through an accessory pathway changes the initial
ventricular activation vector and can generate Q-wave-like
pseudo-infarction patterns without requiring an MI explanation. \\

Revision hypothesis &
Introduce a multi-lead mechanism that replaces purely lead-local
morphology reasoning with cross-lead attention, allowing the
predictor to evaluate axis-level and morphology-level consistency
across limb and precordial leads. \\

Execution and re-evaluation &
The proposed cross-lead revision is implemented as an executable
candidate and evaluated under the fixed data split, metric,
threshold-selection, and leakage constraints. Its behavior is then
replayed on the reviewed case and checked for validation-set
regressions. \\

Replay result &
After replay, the revised predictor suppresses the MI interpretation
and recovers the WPW label. \\

Promotion decision &
The revision is retained only after the executable candidate passes
case replay and regression checks. The resulting trace therefore
connects a concrete failure, raw waveform evidence,
reference-backed measurements, model behavior, and a promoted
model-design update. \\

\bottomrule
\end{tabularx}
\end{table*}

The audit trace in \tabref{tab:app_wpw_case_review} records the
case-specific evidence retained from Evidence-Grounded Failure
Review (EGFR). It is intentionally presented as a development-time
model-design trace rather than as a clinical diagnostic explanation.
The claim is that \sysname uses reproducible measurements produced
through Criteria-to-Measurement Compilation (CMC) to formulate
a testable revision hypothesis, execute it under the fixed problem
contract, and retain it only after replay and regression evaluation;
the trace supports the specific revision on this reviewed case, not a
general guarantee about all WPW cases or all future revisions.

\subsection{Evidence-Grounded Failure Review Example}
\label{app:case_review_latest}

For completeness, \tabref{tab:app_case_review_latest} gives a second,
audited PTB-XL failure-review trace. Unlike the WPW example above, this
trace focuses on a cross-lead hypertrophy error and shows the initial model
scores, measured waveform evidence, candidate revision, replay outcome,
and promotion decision in one development-time record.

\begin{table*}[t]
\centering
\caption{Audited PTB-XL Evidence-Grounded Failure Review example.}
\label{tab:app_case_review_latest}
\small
\setlength{\tabcolsep}{5pt}
\begin{tabularx}{0.95\textwidth}{
    l
    >{\raggedright\arraybackslash}X}
\toprule
Item & Evidence \\
\midrule

Sample &
PTB-XL validation sample 1103. \\

Ground truth &
HYP$=1$, with all other diagnostic superclasses equal to 0. \\

Initial error &
NORM false positive and HYP false negative. \\

Initial model outputs &
NORM score $=0.859$ and HYP score $=0.151$. \\

Raw ECG evidence &
High R-wave amplitude in leads II, aVF, and V4--V6; a deep
S wave in aVR; and an LVH-like cross-lead voltage pattern. \\

Reference-backed measurements &
Lead II maximum $=10.96$, aVF maximum $=10.57$, aVR minimum
$=-9.14$, V4 maximum $=10.34$, V5 maximum $=8.49$, and
V6 maximum $=7.81$. \\

Identified pipeline weakness &
The independent-channel CNN fails to combine the distributed
high-voltage evidence across leads and therefore misses the
cross-lead pattern associated with HYP. \\

Revision hypothesis &
Introduce multi-head cross-lead attention so that the predictor can
jointly evaluate voltage relationships across limb and precordial
leads. \\

Candidate execution &
The proposed revision is implemented and evaluated as an executable
candidate in Iteration 004 under the fixed problem contract. \\

Replay result &
The revised candidate corrects HYP to 1 and suppresses NORM to 0. \\

Score change &
The HYP score increases from $0.151$ to $0.755$, while the NORM
score decreases from $0.859$ to $0.357$. \\

Promotion decision &
The candidate passes the replay and regression checks and is
therefore retained in the final Round 5. \\

\bottomrule
\end{tabularx}
\end{table*}

This case illustrates the intended role of Evidence-Grounded Failure
Review. The review begins with a concrete model failure and jointly
examines the raw ECG, reference-backed measurements produced by
CMC, model outputs, and the relevant cross-lead evidence. It then
formulates a targeted revision hypothesis and admits that hypothesis
to executable candidate testing. The revision is promoted only after
case replay and regression evaluation support its retention.
The LLM is therefore used as an offline model-design controller,
not as an online diagnostic judge.

\subsection{Deployment Efficiency}
\label{app:efficiency_latest}

Efficiency is reported for the frozen predictor produced by the
governed refinement process, rather than for the offline agent search.
The PTB-XL efficiency evaluation shows that \sysname is not the
fastest model in raw inference latency, but it uses few parameters
and FLOPs while achieving higher macro F1 than the strongest
compared baselines.

\begin{table*}[t]
\centering
\caption{Frozen-predictor deployment efficiency on PTB-XL.}
\label{tab:app_efficiency_latest}
\scriptsize
\setlength{\tabcolsep}{3pt}
\resizebox{\textwidth}{!}{%
\begin{tabular}{llrrrrr}
\toprule
Method & Type & Params & FLOPs/sample & ms/sample &
Throughput & Peak mem. MB \\
\midrule
PatchTST
& patch Transformer
& 1.368M & 1.478G & 0.461 & 2,169 & 1,309.7 \\

TimesNet
& frequency-temporal
& 2.505M & 15.620G & 0.573 & 1,746 & 170.0 \\

TS2Vec
& temporal representation
& 0.640M & 1.274G & 0.081 & 12,376 & 188.0 \\

xLSTM-ECG
& ECG-specific xLSTM
& 0.545M & 2.860G & 1.548 & 646 & 4,890.5 \\

\midrule

UniTS
& unified time-series
& 1.066M & 0.903G & 0.236 & 4,244 & 181.8 \\

MERL
& medical ECG representation
& 5.382M & 1.155G & 0.194 & 5,149 & 56.1 \\
\sysname final model
& offline-agent-designed frozen predictor
& 0.487M & 0.086G & 0.116 & 8,601 & 19.2 \\
\bottomrule
\end{tabular}}
\end{table*}

The deployed predictor has higher latency than TS2Vec but lower
latency than most of the compared neural baselines. It also has low
FLOPs and peak memory usage, and it requires no LLM query at
inference time.

Under the default offline development setting, the candidate budget
is $k=3$ and the workflow runs for five iterations. This produces
15 candidate proposals, of which five pass the candidate-level checks
and 10 fail the checks or are rejected. The average agent time is
44.9 minutes per iteration. These quantities characterize one-time model
development cost and should not be interpreted as per-record
diagnostic cost.

\end{document}